\documentclass{amflinkedin}
\usepackage{amsmath,amsfonts,bm}

\def\eqref#1{equation~\ref{#1}}
\def\1{\bm{1}}

\DeclareMathAlphabet{\mathsfit}{\encodingdefault}{\sfdefault}{m}{sl}
\SetMathAlphabet{\mathsfit}{bold}{\encodingdefault}{\sfdefault}{bx}{n}

\newcommand{\E}{\mathbb{E}}

\usepackage{hyperref}
\usepackage{url}
\usepackage{graphicx}
\usepackage{color}
\usepackage{xcolor}         % colors
\usepackage{enumitem}
\usepackage{amsmath}
\usepackage{multirow}
\usepackage{colortbl} 
\usepackage{booktabs}
\usepackage{caption}
\usepackage{siunitx}
\usepackage{amsmath}
\usepackage{amsthm} 
\usepackage{cleveref}
\usepackage{wrapfig}
\usepackage{natbib}
\crefformat{section}{\S#2#1#3}
\crefformat{subsection}{\S#2#1#3}
\crefformat{subsubsection}{\S#2#1#3}
\crefrangeformat{section}{\S#3#1#4 to~\S#5#2#6}
\crefmultiformat{section}{\S#2#1#3}{ and~\S#2#1#3}{, #2#1#3}{ and~#2#1#3}
\Crefformat{figure}{#2Fig.~#1#3}
\Crefmultiformat{figure}{Figs.~#2#1#3}{ and~#2#1#3}{, #2#1#3}{ and~#2#1#3}
\Crefformat{table}{#2Tab.~#1#3}
\Crefmultiformat{table}{Tabs.~#2#1#3}{ and~#2#1#3}{, #2#1#3}{ and~#2#1#3}
\Crefformat{appendix}{#2Appx.~\S#1#3}
\crefformat{algorithm}{Alg.~#2#1#3}
\usepackage{tabularx}
\Crefformat{equation}{#2Eq.~#1#3}

\theoremstyle{plain}
\newtheorem{proposition}{Proposition}
\theoremstyle{definition}

\crefname{proposition}{Prop.}{Props.}
\Crefname{proposition}{Proposition}{Propositions}
\crefname{assumption}{Asm.}{Asms.}
\Crefname{assumption}{Assumption}{Assumptions}

\title{Smaller Models, Better Rejects: \\
Preference Distillation Scaling}

\providecommand{\takeaway}[2]{\noindent\colorbox{gray!12}{%
  \parbox{0.97\linewidth}{\textbf{Takeaway #1.} \emph{#2}}}\par\smallskip}

\author[\clubsuit,\dagger]{Rui Cai\textsuperscript{*}}
\author[\clubsuit, \ddagger]{Wenhui Zhu}
\author[\diamond]{Xiwen Chen}
\author[\clubsuit]{Jincheng Cao}
\author[\clubsuit]{Han Yu}
\author[\clubsuit]{\protect\newline Shayan Mohajer Hamidi}
\author[\triangle]{Zelin He}
\author[\dagger]{Qiyao Ma}
\author[\circ]{Daiwei Chen}
\author[\ddagger]{Xuanzhao Dong}
\author[\clubsuit]{\protect\newline Yuanda Xu}
\author[\clubsuit]{Jelena Markovic-Voronov}
\author[\clubsuit]{Kayhan Behdin}
\author[\clubsuit]{Zhengze Zhou}
\author[\clubsuit]{Ran He}
\author[\clubsuit]{\protect\newline Alborz Geramifard}
\author[\clubsuit]{Rohit Jain}
\author[\dagger]{Zhe Zhao}

\affiliation[\clubsuit]{AI Agentic Modeling and Foundation Team, LinkedIn}
\affiliation[\dagger]{University of California, Davis}
\renewcommand{\affilseparator}{\par}
\affiliation[\ddagger]{Arizona State University}
\renewcommand{\affilseparator}{, }
\affiliation[\diamond]{Clemson University}
\affiliation[\triangle]{Pennsylvania State University}
\renewcommand{\affilseparator}{\par}
\affiliation[\circ]{University of Wisconsin--Madison}
\renewcommand{\affilseparator}{, }

\date{\today}

\correspondence{
  \email{ruicai@ucdavis.edu}
}

\abstract{%
  Preference distillation commonly treats a teacher response as preferred and the student's own response as rejected. This practice rests on two assumptions: the student's own failures are the most informative negatives, and reject responses must come from a model as large as the student, which makes reject generation increasingly costly as students scale. We find that neither holds: across students from 7B to 72B, smaller frozen models generate rejects with less inference compute yet train stronger students than the student's own rejects, both before and after sequence-level knowledge distillation, on code generation and mathematical reasoning.
To analyze this finding, we ask which reject distribution most improves a given student and derive a finite-horizon utility bound in a linearized feature model of Direct Preference Optimization. 
The bound characterizes a favorable region of reject distributions and motivates three interventions.
First, since net transfer in the bound is linear in source mixtures, we randomly mix rejects from a smaller model and from the student-scale model; performance rises as the smaller model's share grows. 
Second, the feature model suggests that task-related content can contribute to reject utility independently of the original prompt pairing, so we reassign rejects to other prompts and further shuffle their code tokens; both still outperform length-matched gibberish, so part of the gain comes from task structure itself. 
Third, the analysis shows that selecting candidates with lower likelihood under the reference policy improves net transfer when higher-likelihood candidates carry less useful contrast. We reselect each source's candidate rejects by this likelihood; lower-likelihood selections outperform higher-likelihood ones for every source. Together, these results suggest a simple design principle: effective rejects preserve task structure while limiting coupling to the reference policy, and smaller frozen models provide both at low cost.
}

\begin{document}

\maketitle
\footnotetext[1]{Work done during internship at LinkedIn.}

\section{Introduction}
\label{sec:introduction}

% Opening and intro to preference distillation
Knowledge distillation transfers the capabilities of a large teacher model to
a smaller student~\citep{hinton2015distilling} and has become a standard
way to build capable large language models at lower
cost~\citep{deepseekr1,zephyr}.
When the teacher is a proprietary model accessible only through its outputs,
black-box knowledge distillation trains a student to imitate the teacher's
responses~\citep{gad,soda}. Beyond imitation, several preference-based distillation methods instead construct supervision from teacher and student responses, treating the teacher output as preferred and the student output as rejected~\citep{plad, dpkd, pad, soda}. SODA applies this idea in a static pipeline with two stages~\citep{soda}. The student first learns from teacher responses through sequence-level knowledge distillation (SeqKD~\citep{seqkd}) and then undergoes DPO~\citep{dpo} on fixed teacher and student responses. Using the student's own response as the reject seems
intuitive because its failures appear to be the most relevant alternatives from which it should learn.

\begin{figure*}[t]
  \centering
  \includegraphics[width=\textwidth]{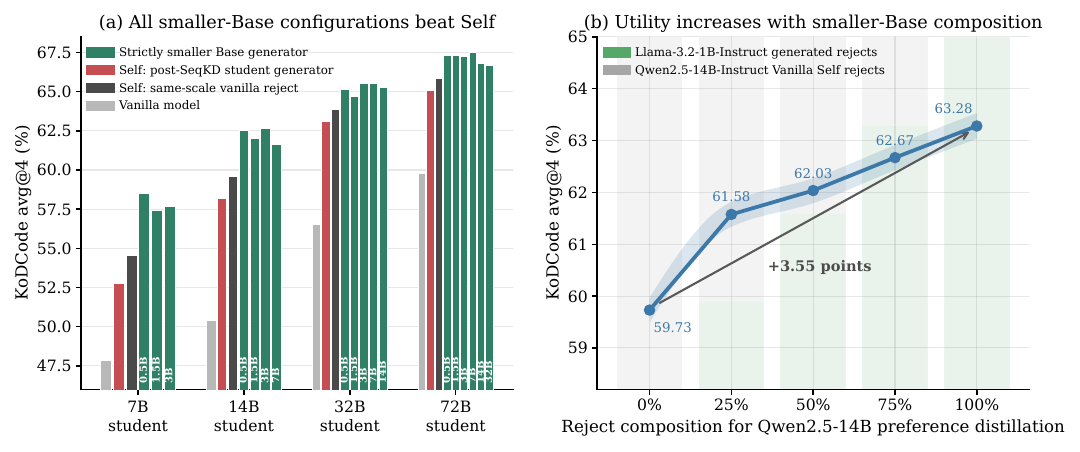}
  \caption{Smaller Base models provide stronger rejects, and their advantage
  varies systematically with source composition. \textit{Left:} KoDCode
  avg@4 for the original vanilla model, both Self constructions, and every
  smaller Base source available to each 7B--72B student. All 18
  smaller Base configurations exceed both post-SeqKD Self and
  vanilla Self. \textit{Right:} a randomized mixture intervention on a matched
  14B population. Avg@4 rises from $59.73$ to $63.28$ as the share of
  Llama-1B Base rejects increases from $0\%$ to $100\%$. }
  \label{fig:intro-preference-scaling}
\end{figure*}

% Problem on reject construction design (with teaser)
This choice raises both a scaling question and a data question. As the student
grows, generating self rejects requires sampling from the same increasingly
large model. Yet the student's own failures, although verified incorrect, may not provide the most useful negative supervision: after SeqKD on teacher responses, they can be near misses that are already likely under the DPO reference, which may leave little for DPO to contrast. \Cref{fig:intro-preference-scaling} summarizes our main
finding. Across Qwen2.5~\cite{qwen25} student models scaling from 7B to 72B, every tested \emph{smaller Base} source, a frozen vanilla model with strictly
fewer parameters than the student, trains a stronger student than either Self
construction: \emph{vanilla Self}, the original model at the student scale,
or \emph{post-SeqKD Self}, the reference policy that initializes DPO. We establish this phenomenon through a large scaling study on verifiable code generation and math reasoning tasks. Across both tasks, smaller Base rejects improve over Self at every student scale, while requiring less inference compute.
% The best reject source still varies across students and tasks: smaller Base sources expand the attainable frontier in both families, while the exact ranking remains target dependent.

% Intro to all experiments and anlysis
Why should the cheaper frozen source also provide stronger negative
supervision? This phenomenon turns reject construction into an inverse design
problem: 
% preference optimization specifies how a chosen reject distribution
% shapes learning, but not which reject distribution provides useful corrective
% supervision
preference optimization determines how the student changes for a given reject distribution, whereas reject construction asks the reverse question of which reject distribution to supply so that the student improves the most~\citep{dpo,did,whatmattersdpo}. 
% We analyze this problem through a
% linearized feature model of DPO, which characterizes reject distributions by their
% transfer to task utility and motivates the construction interventions we study. 
We analyze this problem with a linearized feature model of DPO around the initialization. The model scores each reject source by how much its rejects move training toward better task performance and by what fraction of this movement points the wrong way, and it predicts improvement when the former is large and the latter is small. Applied to three ways of constructing rejects, this theoretical analysis
motivates three interventions on code generation.

First, the analysis shows that the net transfer of a source mixture is linear in its mixing weights, which motivates testing whether utility rises with the
smaller Base share. We randomly mix smaller Base and vanilla Self rejects for a 14B student. Code generation performance
rises from $59.73\%$ to $63.28\%$ as the Llama-1B share grows from $0\%$ to $100\%$ (\Cref{fig:intro-preference-scaling}b), so utility changes
systematically with source composition. 
Second, the analysis shows that reassigning rejects to other prompts, which keeps the same pool of reject
texts but breaks each reject's link to its own prompt, preserves the effect carried by content independent of the prompt. This motivates reassigning
smaller Base rejects to other prompts and further permuting their code tokens. Reassigned rejects outperform length-matched gibberish at every student scale, while gibberish adds little beyond continued training on the chosen responses across all students. Even after permutation destroys token order, syntax, and prompt correspondence, the rejects still outperform gibberish, so part of the gain comes from
task-characteristic code structure rather than from errors specific to each prompt. 
Third, the analysis shows that rejects with lower likelihood under the SeqKD reference help when higher-likelihood rejects offer less useful contrast, and
smaller Base rejects are less likely than Self rejects under this reference. To test this, we sample a fixed pool of candidates from each source for the
14B student and select rejects that favor either lower or higher reference likelihood.
Lower-likelihood selections outperform higher-likelihood ones for every source; for the 1.5B source, avg@4 reaches $63.14\%$ compared with $61.88\%$.
Together, these results suggest that useful rejects preserve task structure while limiting coupling to the reference policy, and smaller Base models
provide both at low cost.

Our contributions are threefold. 
First,
% smaller Base models require less inference compute yet provide stronger rejects than Self across 7B--72B students. 
we establish a robust scaling phenomenon in preference distillation: across 7B--72B students, rejects from smaller frozen Base models require less inference compute yet train stronger students than the student's own rejects.
Second, we cast reject source selection as inverse data design and analyze
it in a linearized feature model of DPO. A finite-horizon bound characterizes a
favorable region of reject distributions and motivates three construction
interventions.
Third, through
these interventions we identify two observable properties of useful rejects,
task-relevant structure and limited coupling to the reference policy, which
yield a simple design principle for reject construction.

\subsection{Related Work}
\label{sec:related-work}

\paragraph{Preference-based distillation.}
Preference-based distillation transfers supervision from stronger models
through comparisons between responses. Existing methods rank responses from
external generators with AI feedback~\citep{zephyr} or treat teacher
responses as preferred to student responses under ranking, distributional,
or listwise objectives~\citep{plad,dpkd,pad}. SODA, closest to our setting,
applies DPO to fixed teacher and student pairs after SeqKD~\citep{soda}.
These methods establish effective forms of preference distillation, but couple the response source to the prescribed training pipeline, while we study the reject-generating distribution as an independent design variable.

\paragraph{Pair selection and optimization.}
Other work selects or weights observed pairs, or couples data generation
with training through online
sampling~\citep{map,difficultyGap,activeDPO,sharpeActive,metaAPO}, or
reshapes how optimization acts on each
pair~\citep{gradientEntanglement,rc-dpo,adaDPO,gateDPO,compassDPO}.
These studies motivate pair-level and optimization-based explanations of data utility. We test such explanations, but the resulting diagnostics do not consistently recover the advantage of strictly smaller Base rejects over Self rejects.

\paragraph{Distribution-level data design.}
DID derives a rejected-response sampling distribution from the differential
information between target and reference policies~\citep{did}; our analysis
instead characterizes useful reject distributions by their effect on task
utility. \citet{cao2026curriculum} show that stronger teachers need not always yield
better students and address this mismatch with a curriculum over progressively
stronger teachers. \citet{whatmattersdpo} find that chosen-response quality often
dominates contrastiveness and on-policy mixing. Source identity and
composition also matter: multi-model pairs can induce superficial shortcuts
in safety alignment~\citep{moreisless}, pairs from weaker generators can
still improve a stronger learner~\citep{deltalearning,decomposingDelta}, and
mixing on-policy and off-policy pairs has task-dependent
effects~\citep{simplemix}. 
Unlike work that changes complete pairs or the chosen-response
source, we hold prompts, chosen responses, initialization, objective,
and training budget fixed while varying reject-source identity, relative
scale, source composition, and reference coupling.

\section{Preference Distillation Setting}
% \label{sec:preliminaries}
% \subsection{Preference Distillation Setting}
\label{sec:prelim-source-effect}

For a student $s$, SeqKD trains on
$\mathcal D_{\mathrm{KD}}=\{(x_i^{\mathrm{KD}},y_i^{\mathrm{KD}})\}$
and produces $S$, which initializes the DPO policy and serves as
its frozen reference. On a disjoint prompt set, reject construction
$q$ produces the preference dataset
\begin{equation}
  \mathcal D_{\mathrm{pref}}(q)
  = \{(x_j^{\mathrm{pref}},y_j^+,y_{j,q}^-)\},
  \label{eq:preference-data}
\end{equation}
where $y_j^+$ is a verified teacher response and $y_{j,q}^-$ is
the reject produced by $q$.
Given a preference triple $(x,y^+,y^-)$, DPO minimizes
\begin{equation}
  \ell_{\mathrm{DPO}}(\theta;x,y^+,y^-)
  =-\log\sigma\!\left(
    \beta\left[
      \log\frac{\pi_\theta(y^+\mid x)}{S(y^+\mid x)}
      -
      \log\frac{\pi_\theta(y^-\mid x)}{S(y^-\mid x)}
    \right]
  \right),
  \label{eq:dpo-loss}
\end{equation}
where $\sigma$ is the sigmoid function and $\beta>0$ controls
the scale of the reference-relative preference margin.
Training minimizes the average loss over
$\mathcal D_{\mathrm{pref}}(q)$.

Let $\operatorname{PD}_s(q)$ denote the policy obtained after
the fixed DPO training protocol. Within each student scale,
all variants share the prompts, chosen responses, SeqKD
initialization, objective, optimizer, training schedule, training
budget, and evaluation protocol. Only the reject construction
changes. For two constructions $q$ and $q'$, we define their
\emph{reject-source effect} as
\begin{equation}
  \Delta_s(q,q')
  =
  \operatorname{Perf}(\operatorname{PD}_s(q))
  -
  \operatorname{Perf}(\operatorname{PD}_s(q')),
  \label{eq:reject-source-effect}
\end{equation}
where $\operatorname{Perf}$ denotes the task performance metric.
This contrast measures the dataset-level effect of replacing
one reject construction with another under the shared protocol.
The following section develops a linearized feature model to analyze how reject composition affects DPO updates and finite-horizon utility.

\section{Reject Source Selection as Inverse Data Design}
\label{sec:theory}

% We study reject construction as an inverse design problem: how should
% we choose the reject distribution to improve a given student? We analyze a policy with a vector of trainable coefficients and derive a sufficient region for improvement over a comparison reject distribution. The region separates how much a source moves the update toward utility improvement from how much it moves the update against it. We then formulate construction hypotheses for source mixtures, structural perturbations, and reference likelihood selection.

We study reject construction as an inverse design problem: how should we
choose the reject distribution to improve a given student? We
analyze DPO on constructed preference pairs through a linearized feature model,
characterize each reject source by two transfer coordinates, and derives three construction interventions from
this characterization, which we evaluate in
\Cref{sec:experiments}. The full construction,
formal statements, and proofs appear in \Cref{sec:app-theory-scope}.

\subsection{A linearized feature model of Reject Utility}
\label{sec:theory-model}

A construction $q$ from \Cref{sec:prelim-source-effect} changes only the
distribution $Q(y^-\mid x)$ of its rejects; the preference-prompt distribution $p$, the distribution
$T(y^+\mid x)$ of verified teacher responses, and the SeqKD reference $S$
are shared across constructions. Let $\phi(x,y)\in\mathbb R^d$ be fixed
response features and $v\in\mathbb R^d$ a trainable coefficient vector, and
consider
\begin{equation}
  \pi_v(y\mid x)=
  \frac{S(y\mid x)\exp\!\left(v^\top\phi(x,y)\right)}
       {\mathbb E_{y'\sim S(\cdot\mid x)}
        \exp\!\left(v^\top\phi(x,y')\right)},
  \qquad \pi_0=S.
  \label{eq:theory-feature-policy}
\end{equation}
Taking $\phi(x,y)=\nabla_\theta\log\pi_\theta(y\mid x)\big|_{\theta_0}$,
the score gradients of the reference network at its SeqKD parameters
$\theta_0$, makes $\pi_v$ the network's first-order model around the shared
DPO initialization,
following the local linearization perspective on neural
training~\citep{linearizedNetworks} (\Cref{sec:app-vector-model}). In this
geometry a reject can share features with the chosen response and contrast
along others~\citep{gradientEntanglement,likelihoodDisplacement}: for a task
requiring \texttt{max(xs)}, the reject \texttt{min(xs)} contrasts on maximum
versus minimum selection, whereas \texttt{max(0, max(xs))} shares maximum
selection and differs only in an erroneous zero lower bound
(detailed illustration in \Cref{sec:app-theory-code-example}).

Let $\delta=\phi(x,y^+)-\phi(x,y^-)$, and write $\mathbb E_Q[\cdot]$ for the
expectation over triples
$(x,y^+,y^-)\sim p(x)\,T(y^+\mid x)\,Q(y^-\mid x)$. The shared normalizer
cancels, so a pair's DPO margin in \Cref{eq:dpo-loss} is
$\beta v^\top\delta$, and full-batch gradient descent with step sizes
$\eta_t>0$ follows
\begin{equation}
  v_{t+1}^Q=v_t^Q+\eta_t F_Q(v_t^Q),\qquad
  F_Q(v)=\beta\,\mathbb E_Q\!\left[
    \sigma(-\beta v^\top\delta)\,\delta\right],\qquad v_0^Q=0.
  \label{eq:theory-feature-field}
\end{equation}
Each pair pushes $v$ along $\delta$ with a weight that decays as its
margin grows; the construction enters the dynamics only through the
distribution of $\delta$.

\subsection{Feature Transfer and a Favorable Region}
\label{sec:theory-region}

\Cref{eq:theory-feature-field} specifies how a construction moves $v$; we
now ask how that movement translates into task performance, and for which
reject distributions it helps. We measure utility by expected task reward,
denoted as $U(v)=\mathbb E_{x\sim p_{\mathrm{eval}},y\sim\pi_v(\cdot\mid x)}[r(x,y)]$,
where $p_{\mathrm{eval}}$ is the evaluation-prompt distribution and the reward
$r(x,y)\in[0,1]$ scores task success, and let $u=\nabla U(0)\ne0$ and
$\widehat u=u/\|u\|$, the direction in which an initial update most improves
utility. A reject is useful or harmful only relative to what it replaces,
so we fix a comparison reject distribution $Q_0(y\mid x)$ with the same $p$
and $T$. Let
$\bar\phi_0(x)=\mathbb E_{y\sim Q_0(\cdot\mid x)}\phi(x,y)$, and score a
reject by the transfer score
\begin{equation}
  k(x,y)=\left\langle\widehat u,\,\bar\phi_0(x)-\phi(x,y)\right\rangle,
  \label{eq:theory-transfer-score}
\end{equation}
positive when substituting $y$ for the comparison reject moves the initial
update further along the utility direction. Two coordinates, and the net transfer they imply, summarize a source:
\begin{equation}
  a_Q=\mathbb E_{x\sim p,\,y\sim Q}|k(x,y)|,\qquad
  c_Q=\frac{\mathbb E_{x\sim p,\,y\sim Q}[-k(x,y)]_+}{a_Q},\qquad
  \kappa_Q=\mathbb E\,k=a_Q(1-2c_Q),
  \label{eq:theory-transfer-coordinates}
\end{equation}
with $[w]_+=\max(w,0)$, all expectations over $x\sim p$ and $y\sim Q$, and
$c_Q=0$ when $a_Q=0$. Here $a_Q$ is the transfer mass (how much the source
moves the update along $\widehat u$ at all) and $c_Q$ the adverse share
(the fraction of that movement pointing the wrong way), so $\kappa_Q$ is
the net transfer that survives cancellation.
Because prompts and chosen responses are shared,
the initial field difference obeys the exact identity
$\langle u,F_Q(0)-F_{Q_0}(0)\rangle=\frac{\beta}{2}\|u\|\kappa_Q$. Over a
finite horizon, this initial advantage carries through training:

\begin{proposition}[Finite-horizon transfer bound]
\label{prop:theory-gain}
Under regularity conditions shared by the compared constructions (bounded
features suffice; \Cref{sec:app-theory-proofs}), $H$ steps with positive
step sizes $\eta_t$ and total step size $\tau_H=\sum_{t=0}^{H-1}\eta_t$
give
\begin{equation}
  G_H(Q)=U(v_H^Q)-U(v_H^{Q_0})
  \;\ge\; K_H\,a_Q(1-2c_Q)-E_H,
  \qquad K_H=\frac{\beta\tau_H\|u\|}{2},
  \label{eq:theory-gain-bound}
\end{equation}
where $E_H=O(\tau_H^2)$ collects field drift and utility curvature and is
given explicitly in \Cref{sec:app-theory-proofs}.
\end{proposition}

For a target gain
$\epsilon>0$, the bound implies $G_H(Q)\ge\epsilon$ on the favorable
region
\begin{equation}
  \mathcal F_{H,\epsilon}
  =\left\{Q:
    0\le c_Q<\frac12,\quad
    a_Q\ge\frac{\epsilon+E_H}{K_H(1-2c_Q)}
    \right\}.
  \label{eq:theory-favorable-region}
\end{equation}
% The two requirements trade off, as a
% larger adverse share demands stronger transfer, and because $K_H$ is linear
% in the step-size schedule while $E_H$ is quadratic, any source with
% $\kappa_Q>0$ enters the region once the budget is small enough. The
% coordinates characterize a reject distribution at initialization; $H$
% bounds the range over which that characterization stays informative.
The two conditions trade off: a larger adverse share $c_Q$ requires a larger transfer mass $a_Q$. Because $K_H$ is linear in the total step size $\tau_H$ while $E_H=O(\tau_H^2)$, scaling all step sizes by a sufficiently small factor makes the bound in \Cref{eq:theory-gain-bound} positive for any source with $\kappa_Q>0$, and membership in $\mathcal F_{H,\epsilon}$ is exactly the  condition that this bound is at least $\epsilon$. 

\subsection{Construction Interventions and Hypotheses}
\label{sec:theory-interventions}

The coordinates $a_Q$ and $c_Q$ are not directly observable. We therefore
translate them into three construction operators, each of which moves
$\kappa_Q$ in a direction fixed by an explicit hypothesis about the source,
and state the hypotheses as P1--P3. For a construction $q$ inducing $Q$, an
operator $\mathcal O$ yields a construction $q_{\mathcal O}$ inducing
$\mathcal OQ$, whose effect is the causal contrast
$\Delta_s(q_{\mathcal O},q)$ of \Cref{eq:reject-source-effect}.

\paragraph{H1: source mixtures.}
Randomized source assignment implements
\begin{equation}
  \mathcal M_{\mathbf w}(Q_1,\ldots,Q_m)=\sum_{s=1}^m w_sQ_s,
  \qquad \mathbf w\in\Delta^{m-1},
  \label{eq:theory-mixture}
\end{equation}
with $\Delta^{m-1}$ the probability simplex. Because $k$ is fixed by the
common reference, utility, and comparison distribution,
$\kappa_{\mathcal M_{\mathbf w}}=\sum_s w_s\kappa_{Q_s}$ and
$a_{\mathcal M_{\mathbf w}}=\sum_s w_sa_{Q_s}$: net transfer is affine in
$\mathbf w$, which motivates testing whether utility rises along the mixture path as
the smaller Base share grows (\Cref{sec:exp-source-effect}).

\paragraph{H2: prompt reassignment and lexical permutation.}
With $\overline Q(y)=\mathbb E_{x'\sim p}Q(y\mid x')$ the prompt-averaged
reject distribution, reassignment applies
\begin{equation}
  (\mathcal P_\rho Q)(y\mid x)=(1-\rho)Q(y\mid x)+\rho\,\overline Q(y).
  \label{eq:theory-reassignment}
\end{equation}
For an additive decomposition
$\phi=\phi_{\mathrm{dom}}(y)+\phi_{\mathrm{pair}}(x,y)$, the mean of
$\phi_{\mathrm{dom}}$ is invariant under $\mathcal P_1$
(\Cref{sec:app-theory-selection}), so the transfer it carries survives,
and lexical permutation removes all but its order-invariant part. This
motivates testing whether both retain utility over length-matched gibberish
(\Cref{sec:exp-localization}).

% We use length-normalized likelihood under the SeqKD reference as an
% operational measure of source--reference coupling, and use ``reference
% coupling'' as shorthand below.

\paragraph{H3: reference-likelihood reselection.}
Within a generator-specific candidate bank $C_x$, let $\tilde s_j$
standardize the length-normalized reference score
$s_j=|y_j|^{-1}\log S(y_j\mid x)$ within the bank. The reselection rule samples
\begin{equation}
  q_\gamma(j\mid C_x)=\frac{\exp(-\gamma\tilde s_j)}
  {\sum_{l}\exp(-\gamma\tilde s_l)},
  \label{eq:theory-reference-tilt}
\end{equation}
inducing the reject distribution $Q_\gamma$, with $\gamma>0$ repelling from the
reference, $\gamma<0$ attracting, and $\gamma=0$ native. Then
$d\kappa_{Q_\gamma}/d\gamma=-\mathbb E_{x\sim p}
\operatorname{Cov}_{j\sim q_\gamma}(\tilde s_j,k)$
(\Cref{sec:app-theory-selection}), so repulsion raises net transfer whenever
candidates with higher reference likelihood transfer less. The construction
hypothesis is that this holds within model-generated candidate banks: near
misses that share features with the correct response sit closer to the
reference yet supply less contrast, as in the two-feature example of
\Cref{sec:app-theory-code-example}. This motivates testing whether repulsion
outperforms attraction within fixed candidate banks
(\Cref{sec:exp-mechanism}).

\section{Experiments}
\label{sec:experiments}

\subsection{Experimental Settings}
\label{sec:exp-setup}

We study the two stage preference-distillation pipeline introduced by SODA~\citep{soda}. A Qwen2.5 student first learns execution verified teacher responses through SeqKD and then undergoes DPO on a disjoint preference set. Within each student scale, all DPO variants share the SeqKD initialization, prompts, chosen responses, objective, optimization budget, and evaluation protocol. Only reject construction changes. A \emph{smaller Base source} is a frozen vanilla model with strictly fewer parameters than the student. We compare these sources with vanilla Self and post-SeqKD Self. Vanilla Self uses the matching instruction tuned model, while post-SeqKD Self uses the policy that initializes DPO. We write Self for both when the distinction is immaterial. 
Our main scaling study covers Qwen2.5-Instruct students at 7B, 14B, 32B, and 72B~\citep{qwen25}. The code line uses KoDCode~\citep{kodcode}. SeqKD trains on  75,752 prompts with execution verified GPT-4o responses~\citep{gpt4o}. DPO uses a disjoint preference split of 24,248
prompts. Each prompt has a separately generated and verified GPT-4o response as chosen, together with a reject from the designated frozen source. Every reject is verified incorrect: it fails at least one unit test. We evaluate on 5,000 held out KoDCode problems and four external benchmarks:
BigCodeBench-Complete, BigCodeBench-Instruct~\citep{bcb},
HumanEval~\citep{humaneval}, and MBPP~\citep{mbpp}. The primary in domain
endpoint is avg@4, the mean execution success over four sampled responses.
The mathematical reasoning line uses verifier correct DeepSeek-R1 trajectories
from OpenR1-Math-220k~\citep{deepseekr1,openr1math}. These trajectories provide
the teacher responses for SeqKD and the chosen responses for preference training. Every reject is sampled from the designated source and verified incorrect by the answer verifier. We evaluate avg@4 on MATH-500~\citep{math500} and the 2024 and 2025
AIME problems~\citep{aime}. Complete split
construction, training, decoding, and evaluation details in
\Cref{sec:app-experimental-details}.

% \begin{figure*}[t]
%   \centering
%   \includegraphics[width=\textwidth]{sections/figures/subscale_reject_strategy.pdf}
%   \caption{Absolute performance across distillation stages and reject
%   constructions. Panels \textbf{(a)} through \textbf{(d)} show the 7B, 14B, 32B,
%   and 72B students. Each panel reports KoDCode held-out avg@4 and the
%   problem-count-weighted external pass@1 average. Bars include Base, SeqKD,
%   \textit{Continued-SFT}, vanilla Self, post-SeqKD Self, and every strict-subscale Base
%   reject available for that student.}
%   \label{fig:subscale-reject-strategy}
% \end{figure*}

\begin{figure*}[t]
  \centering
  \includegraphics[width=\textwidth]{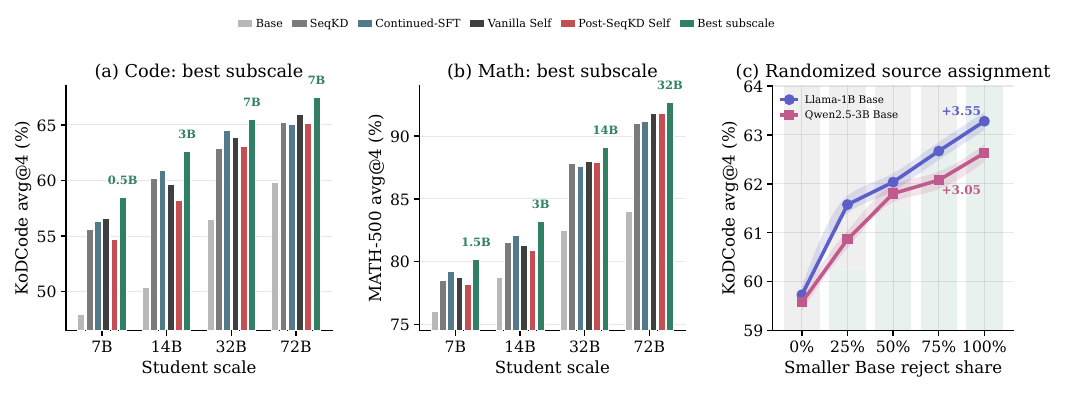}
  \caption{Smaller Base models improve preference distillation across code
  generation and mathematical reasoning. \textbf{(a, b)} Absolute avg@4 for
  the original vanilla model, SeqKD, Continued-SFT, both Self constructions, and
  the best observed smaller Base source for each 7B--72B student.
  Labels above the green bars identify the displayed reject source.
  \textbf{(c)} Two randomized source assignment interventions vary the
  proportions of smaller Base and vanilla Self rejects for a 14B student. The green and gray
  background shows the dataset composition. Performance increases with the
  smaller Base share for both source families.}
  \label{fig:preference-scaling-mixture-result}
\end{figure*}

\subsection{RQ1: How Should Reject Generation Scale with the Student?}
\label{sec:exp-source-effect}

\paragraph{Smaller Base sources consistently outperform Self.}
\Cref{fig:preference-scaling-mixture-result} summarizes the best observed smaller Base endpoint at every student scale on both code generation and
mathematical reasoning. 
Across the complete code scaling matrix, all smaller Base configurations outperform both Self
constructions on avg@4 (\Cref{tab:kodcode_heldout_raw}). The gains are
substantial across both task families. For example, rejects generated by
Qwen2.5-3B-Instruct improve the 14B student's avg@4 over vanilla Self by
$3.1$ points on code generation and $1.9$ points on mathematical reasoning.
The best source varies with the student and task. Llama Base sources also remain
competitive with the best Qwen2.5 smaller Base source
(\Cref{sec:app-cross-family}). Complete
in-domain and external code results are reported
in \Cref{tab:kodcode_heldout_raw,tab:kodcode_external_pass1}, and the complete
mathematical reasoning results are reported in
\Cref{tab:math500_aime_avg4}.  These sources are also cheaper: on the code preference prompts, we estimate generation compute using
parameter count adjusted by output length. For smaller Base sources,
this estimate is $0.9\%$ to $50.1\%$ of vanilla Self at the same student
scale, corresponding to a $2.0\times$ to $108.3\times$ reduction. The complete
cost analysis and measured GPU runtimes appear in
\Cref{sec:app-reject-generation-cost}.

\paragraph{Randomized source assignment controls performance.}
The pure source comparison changes an entire reject dataset at once.  We therefore apply the randomized source assignment operator of
\Cref{eq:theory-mixture} (H1 in \Cref{sec:theory-interventions}), varying the proportions of smaller Base and vanilla Self rejects on a matched 14B training population. As shown in \Cref{fig:preference-scaling-mixture-result}c, increasing the share of Llama-1B
Base rejects from $0\%$ to $100\%$ raises code
avg@4 from $59.73\%$ to $63.28\%$. Utility therefore changes systematically with source composition rather than only between the two pure endpoints. The
Qwen2.5-3B intervention follows the same ordering from $59.58\%$ to
$62.63\%$, and a 72B midpoint lies between its two pure source endpoints. Full composition results are reported in
\Cref{sec:app-mixture-results}.

\takeaway{1}{Reject generation need not scale with the student. Across the
tested 7B to 72B range, smaller Base models use less inference
compute and provide stronger rejects than either Self construction.}

\subsection{RQ2: Where Does the Smaller Base Gain Reside?}
\label{sec:exp-localization}

The scaling study identifies a stable source effect, but it does not determine
whether the gain requires negatives at all, whether any suppressible text is
sufficient, or whether the gain depends on the exact prompt and reject pairing.
We address these possibilities with direct construction interventions, realizing the two operators of H2
(\Cref{sec:theory-interventions}), first
localizing the gain to the reject source distribution and then identifying a
structural component carried by that distribution. 

\begin{figure*}[t]
  \centering
  \includegraphics[width=\textwidth]{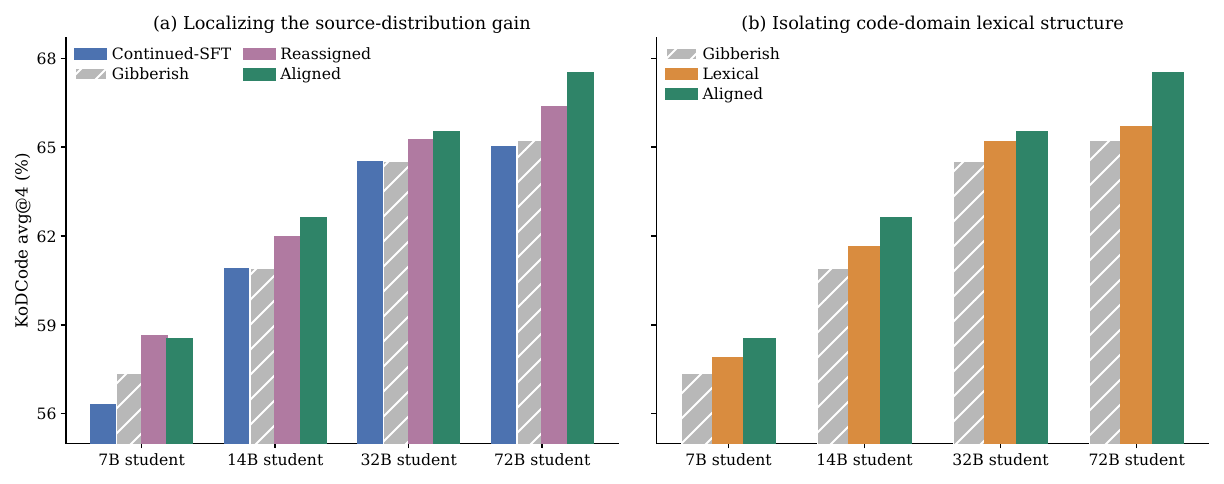}
  \caption{Localizing and decomposing the utility of smaller Base rejects.
  \textbf{(a)} Absolute KoDCode avg@4 across four control settings at every
  student scale. Continued-SFT trains on the preference-stage chosen responses
  alone. Gibberish supplies length-matched artificial rejects. Prompt
  reassignment preserves real Base rejects while removing their original
  prompt correspondence, and Aligned restores that correspondence.
  \textbf{(b)} Lexical permutation preserves the generated code-token
  inventory of prompt-reassigned Base rejects while destroying token order,
  syntax, and coherent program semantics. Aligned shows the original
  prompt-matched Base construction.}
  \label{fig:localization-structure}
\end{figure*}

\paragraph{Real source distributions retain utility without prompt
correspondence.}
We construct four controls at every student scale using the strongest in
domain Base source for that student. Continued-SFT starts from the SeqKD model
and trains on the DPO-stage chosen responses alone. Gibberish replaces
every reject with length-matched artificial text, serving as the comparison
construction for H2. Prompt reassignment realizes $\mathcal P_1$ in \Cref{eq:theory-reassignment} by assigning real Base rejects to different prompts, with interface normalization and execution filtering. The aligned setting restores the original prompt and reject
correspondence. As shown in \Cref{fig:localization-structure}a, Gibberish closely
tracks Continued-SFT at 14B, 32B, and 72B, with a larger gain at 7B.
Prompt-reassigned Base rejects improve over Gibberish at every scale. The largest gain appears for the
7B student, where reassignment improves avg@4 over Gibberish by $1.28$
points.Together they show that part of the gain does not require the original prompt and reject pairing and can come from features of the reject text alone; the incremental effect of restoring alignment depends on the student and source. Complete endpoints, paired contrasts, and reward
trajectories appear in \Cref{sec:app-localization-details}.

\paragraph{Code-domain lexical structure carries corrective value.}
We next permute the lexical tokens within every prompt-reassigned Base
reject, keeping only the token inventory, the order-invariant part of the text feature in H2.
This operation preserves the generated code-token inventory while destroying
token order, Python syntax, coherent programs, and prompt correspondence.
Despite this destruction, Lexical consistently outperforms Gibberish at every
student scale, as shown in \Cref{fig:localization-structure}b. The largest gain
occurs for the 14B student, where preserving only the generated code-token
inventory improves avg@4 by $0.73$ points. Code-domain lexical structure therefore carries corrective value even without
an executable program, coherent failure semantics, or prompt correspondence. An AST-based
intervention and complete paired contrasts are reported in
\Cref{sec:app-structure-details}.

\takeaway{2}{The utility of real Base rejects survives removal of exact prompt
correspondence, and code-domain lexical structure recovers a substantial part
of this distribution-level gain across every student scale.}

\subsection{RQ3: What Makes a Reject Distribution Useful?}
% \subsection{RQ3: Why Do Smaller Base Rejects Outperform Self Rejects?}
\label{sec:exp-mechanism}

RQ2 shows that useful rejects retain structure from the code domain. We
further analyze what distinguishes smaller Base sources from Self within this
structured output family. The analysis in \Cref{sec:theory} identifies
likelihood under the SeqKD reference as a property of the reject distribution
that we can manipulate directly, and H3 in \Cref{sec:theory-interventions}
predicts an advantage for repulsion over attraction when higher-likelihood
candidates transfer less. We test this in two steps. First, we ask whether
reference likelihood organizes the utility of naturally generated reject
sources across student scales. Second, we manipulate this quantity directly by
reselecting from fixed candidate banks.

\begin{figure*}[t]
  \centering
  \includegraphics[width=\textwidth]{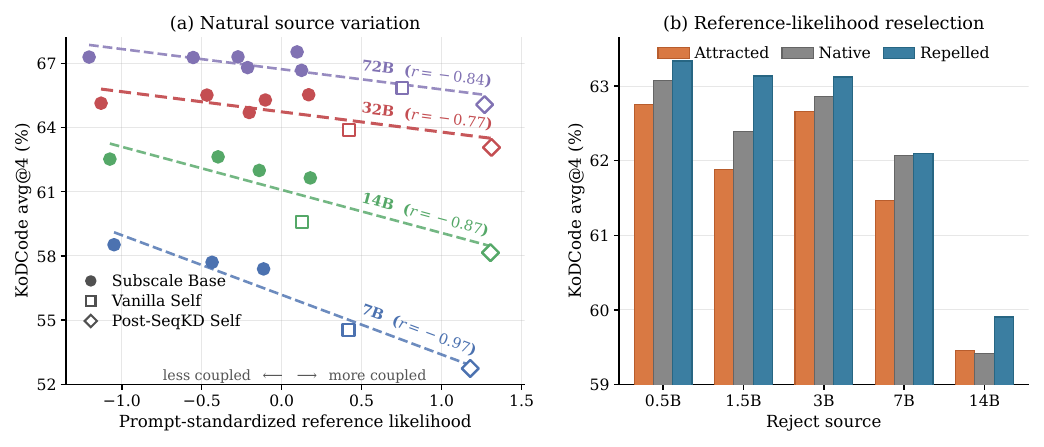}
  \caption{Reference likelihood tracks and controls reject utility.
  \textbf{(a)} Each point is a natural reject source scored by its mean
  length-normalized likelihood under the corresponding student's SeqKD
  reference and standardized within prompt. Lines are fitted separately for
  each student scale; labels report Pearson correlations. Filled circles are
  smaller Base sources, squares are vanilla Self, and diamonds are
  post-SeqKD Self. \textbf{(b)} For a fixed Qwen2.5-14B-Instruct student, each
  source-specific candidate bank is reselected toward lower reference
  likelihood (Repelled), without likelihood preference (Native), or toward
  higher reference likelihood (Attracted).}
  \label{fig:reference-coupling-evidence}
\end{figure*}

\paragraph{Reference likelihood organizes the natural source boundary.}
For every student, we score each reject source under that student's exact
SeqKD reference on the same preference prompts. As shown in
\Cref{fig:reference-coupling-evidence}a, higher reference likelihood is associated with lower downstream avg@4 at every student scale.
The strongest relationship occurs for the 7B student, with a Pearson
correlation of $-0.97$. Smaller Base sources consistently combine lower coupling with higher utility, while vanilla Self and post-SeqKD Self combine higher coupling with lower utility. Reference likelihood thus
organizes the stable source boundary between smaller Base and Self rejects.
Complete source measurements and robustness analyses are
reported in \Cref{sec:app-natural-coupling}.

\paragraph{Lower reference coupling improves utility under controlled
reselection.}
To separate coupling from the choice of generator, we fix each reject source and change only which of its responses we use. For each prompt, the source generates eight candidates, and we score each one by its per-token log likelihood under the 14B student's SeqKD reference. From these candidates we pick one reject in three ways (\Cref{eq:theory-reference-tilt}): Repelled favors low likelihood, Native picks at random, and Attracted favors high likelihood. 
A prompt is then kept only if all three selected rejects run to completion, pass exactly the same fraction of unit tests, and differ in length by at most 32 tokens. 
The generator, candidate bank, student
initialization, and DPO protocol remain fixed within each source.
\Cref{fig:reference-coupling-evidence}b shows that lower reference likelihood improves utility in our candidate banks:
Repelled produces the strongest endpoint for every reject source, while Attracted selection reduces utility for each smaller Base source. The largest
separation occurs for the 1.5B source, where Repelled reaches $63.14\%$
avg@4 compared with $61.88\%$ for Attracted, a gain of $1.26$ points.
Because reselection changes reference likelihood without changing the
generator or its candidate bank, this result supports treating reference coupling as
an operational property of the reject distribution rather than only a proxy for
model scale. Manipulation
checks, raw endpoints, paired effects, and interaction contrasts for this intervention are reported in \Cref{sec:app-coupling-reselection}.

Together, RQ2 and RQ3 suggest why smaller Base sources are effective.
Gibberish lacks the code-domain structure that carries corrective value. Self
rejects preserve this structure but remain strongly coupled to the policy used
as the DPO reference. Smaller Base sources have both properties:
they retain code-domain structure while remaining less coupled to the
reference. Reference repulsion lowers coupling further and improves utility.
% Additionally, we evaluate aggregate reject fields, implicit reward
% margins, fixed-pair likelihood trajectories, and an RC-DPO intervention.
% These analyses describe individual pairs or optimization behavior, but do not recover the stable smaller Base over Self ordering. We report them in \Cref{sec:app-explanation-boundaries} as complementary evidence delimiting our interpretation.
Additionally, optimization diagnostics, including reject fields, reward margins, likelihood trajectories, and RC-DPO, do not recover the smaller Base over Self ordering (\Cref{sec:app-explanation-boundaries}).

\takeaway{3}{Across natural reject sources, lower reference coupling
separates smaller Base from Self rejects. Within fixed source-specific
candidate banks, reference repulsion improves downstream utility. Task-related
structure and limited reference coupling are the two theory-motivated source
properties that we identify empirically.}

\newpage
\section{Conclusion}
\label{sec:conclusion}

In this paper, we revisit a common default in preference distillation: using
the student's own failures as rejects. Across students from 7B to 72B, smaller
frozen Base models generate reject responses with less inference compute yet train
stronger than both Self constructions across code generation and math reasoning tasks. Reject generation therefore need not scale with the student.

To analyze this finding, we cast reject source selection as inverse data
design and derive a finite-horizon bound in a linearized feature model of DPO,
which motivates three interventions. Performance rises as smaller Base rejects
replace student-scale rejects in a randomized mixture. Length-matched
gibberish provides no residual benefit beyond continued training on chosen
responses, whereas rejects reassigned to other prompts, and even rejects that
keep only their code tokens, outperform it, so part of the gain comes from
task structure rather than from errors specific to each prompt. Selecting
candidates with lower reference likelihood outperforms selecting those with
higher likelihood for every source. Together, these results suggest that
effective rejects preserve task structure while limiting coupling to the
reference policy, and smaller frozen models provide both at low cost.

\newpage
\bibliography{sections/iclr2027_conference}
\bibliographystyle{styles/iclr2027_conference}

\clearpage
\appendix
% ---------------------------------------------------------------------
% Appendix: complete numerical results and supporting protocols.
% Included by main.tex after the bibliography.
% ---------------------------------------------------------------------

\appendix
\crefalias{section}{appendix}
\crefname{appendix}{appendix}{appendices}
\Crefname{appendix}{Appendix}{Appendices}

\section{Experimental Details}
\label{sec:app-experimental-details}

Our main scaling study covers Qwen2.5 students at 7B, 14B, 32B, and 72B. We
train and evaluate separate lines for verifiable code generation and
mathematical reasoning. 

\textbf{Code training line.}
The main scaling experiment uses KoDCode~\citep{kodcode}. SeqKD trains on
75,752 prompts with execution-verified teacher solutions. The
preference stage uses a disjoint set of 24,248 prompts, each paired with a
separately generated, execution-verified teacher response $y^+$ and a reject $y_q^-$ from the designated frozen source. Each reject is verified incorrect by execution: it fails at least one unit test. All Continued-SFT and DPO variants
within a student scale use the same preference prompts and $y^+$ responses;
only DPO uses rejects, and only the reject source changes across DPO variants.

\textbf{Math training line.}
We use a separate OpenR1-Math line to test whether residual value beyond
Continued-SFT also appears in mathematical reasoning. Students are first
trained on filtered verifier-correct reasoning trajectories (33,524 prompts for SeqKD) and then receive
Continued-SFT or preference training on OpenR1-Math preferences (12,288 prompts for DPO). Each reject is sampled from the designated source and retained only if the automatic answer verifier marks its final answer incorrect. 

\textbf{Code evaluation.}
The in-domain test contains 5,000 KoDCode problems disjoint from both training
stages. We report greedy pass@1 and four stochastic responses per problem:
avg@4 is the mean execution success rate, and pass@4 is the fraction of
problems solved at least once. For KoDCode-trained models, external evaluation
uses BigCodeBench-Complete (BCB-C) and BigCodeBench-Instruct
(BCB-I)~\citep{bcb}, which contain the same 1,140 problems under different
prompt formats, together with HumanEval~\citep{humaneval} (164 problems) and
MBPP~\citep{mbpp} (257 problems). The reported external average weights each
benchmark by its number of problems and is a descriptive summary.

\textbf{Math evaluation.}
OpenR1-Math-trained models are evaluated on external benchmarks: MATH-500~\citep{math500} and AIME~\citep{aime}
using an automatic answer verifier. We report avg@4 over four sampled
responses.

\textbf{Training hyperparameters.}
\Cref{tab:app-hyperparameters} lists the training settings of the code line.
Within each student scale, all Continued-SFT and DPO variants use the same
settings; only the training objective and the reject source differ.

\begin{table}[h]
\centering
\small
\caption{Training hyperparameters of the code line. All runs use
full-parameter training with AdamW, a cosine learning rate schedule with 10\%
linear warmup, bf16 precision, gradient checkpointing, DeepSpeed ZeRO-3. Continued-SFT uses the same settings as DPO.}
\label{tab:app-hyperparameters}
\begin{tabular}{lcc}
\toprule
Setting & SeqKD & DPO \\
\midrule
Learning rate & $5\times10^{-6}$ & $5\times10^{-7}$ \\
DPO $\beta$ & -- & 0.1 \\
Epochs & 1 & 1 \\
Gradient accumulation & 16 & 16 \\
Effective batch (7B--72B) & 256 & 256 \\
Max sequence length & 2,560 & 2,560 \\
Gradient clipping & 0.2 & 0.2 \\
\bottomrule
\end{tabular}
\end{table}

\clearpage
\section{Reject Generation Cost Analysis}
\label{sec:app-reject-generation-cost}

We measure reject-generation cost on the exact 24,248 selected responses used
by the code preference stage. Let $P_q$ denote the number of unique parameters
in reject generator $q$, $N_x$ the total number of prompt tokens, and $N_q^-$
the total number of generated reject tokens. We define the decode and total
parameter-token proxies as
\begin{equation}
  C_{\mathrm{decode}}(q)=P_qN_q^-,
  \qquad
  C_{\mathrm{total}}(q)=P_q(N_x+N_q^-).
  \label{eq:app-reject-generation-cost}
\end{equation}
The first quantity measures autoregressive output work, while the second also
charges prompt prefill to each generator. Both account for source-dependent
response length and are reported as compute proxies rather than measured
FLOPs. The shared prompts contain 9.536 million Qwen2.5 tokens, or 393.3
tokens per prompt on average.

Parameter count alone is insufficient because reject lengths differ across
sources: the 0.5B source averages 466.7 tokens per reject, while the other
Qwen2.5 sources average 240 to 305 tokens. The proxies in
\Cref{tab:reject-generation-cost-comparison} account for this difference, and
each cost is normalized by the same-scale vanilla Self generator used for the
corresponding student.

\begin{table*}[t]
  \centering
  \caption{Complete output-length-adjusted parameter-token cost comparison.
  Decode and total costs are percentages of same-scale vanilla Self. The
  reduction factor is the inverse of the total-cost ratio.}
  \label{tab:reject-generation-cost-comparison}
  \scriptsize
  \begin{tabular}{llrrrrr}
    \toprule
    Student & Smaller Base source & Parameter ratio & Reject tokens
      & Decode cost & Total cost & Reduction \\
    \midrule
    7B  & 0.5B & 6.5\%  & 11.317M & 10.6\% & 8.2\%  & $12.2\times$ \\
    7B  & 1.5B & 20.3\% &  6.366M & 18.7\% & 19.6\% & $5.1\times$ \\
    7B  & 3B   & 40.5\% &  7.074M & 41.5\% & 40.9\% & $2.4\times$ \\
    \midrule
    14B & 0.5B & 3.3\%  & 11.317M & 5.1\%  & 4.1\%  & $24.3\times$ \\
    14B & 1.5B & 10.5\% &  6.366M & 9.0\%  & 9.8\%  & $10.2\times$ \\
    14B & 3B   & 20.9\% &  7.074M & 20.0\% & 20.5\% & $4.9\times$ \\
    14B & 7B   & 51.6\% &  6.908M & 48.2\% & 50.1\% & $2.0\times$ \\
    \midrule
    32B & 0.5B & 1.5\%  & 11.317M & 2.6\%  & 2.0\%  & $50.9\times$ \\
    32B & 1.5B & 4.7\%  &  6.366M & 4.6\%  & 4.7\%  & $21.3\times$ \\
    32B & 3B   & 9.4\%  &  7.074M & 10.3\% & 9.8\%  & $10.2\times$ \\
    32B & 7B   & 23.2\% &  6.908M & 24.9\% & 23.9\% & $4.2\times$ \\
    32B & 14B  & 45.1\% &  7.387M & 51.6\% & 47.7\% & $2.1\times$ \\
    \midrule
    72B & 0.5B & 0.7\%  & 11.317M & 1.3\%  & 0.9\%  & $108.3\times$ \\
    72B & 1.5B & 2.1\%  &  6.366M & 2.3\%  & 2.2\%  & $45.5\times$ \\
    72B & 3B   & 4.2\%  &  7.074M & 5.2\%  & 4.6\%  & $21.8\times$ \\
    72B & 7B   & 10.5\% &  6.908M & 12.5\% & 11.2\% & $8.9\times$ \\
    72B & 14B  & 20.3\% &  7.387M & 25.8\% & 22.4\% & $4.5\times$ \\
    72B & 32B  & 45.1\% &  6.460M & 50.1\% & 47.0\% & $2.1\times$ \\
    \bottomrule
  \end{tabular}
\end{table*}

Across all strict-subscale configurations, the total-cost proxy ranges from
$0.9\%$ to $50.1\%$ of same-scale vanilla Self, corresponding to reductions
of $2.0\times$ to $108.3\times$. Relative to post-SeqKD Self, the range is
$0.9\%$ to $55.1\%$, corresponding to reductions of $1.8\times$ to
$108.3\times$.

We additionally calibrate the proxy with a matched H100 generation sweep in
which each source generates eight candidates for the same 24,248 prompts with
vLLM, temperature $0.6$, and a 2,048-token output cap. Counting only the GPUs
used for tensor-parallel inference, the 0.5B, 1.5B, 3B, and 7B sources
require $1.22$, $0.53$, $0.64$, and $0.84$ H100 hours, compared with $2.37$
for the 14B generator. The measured runtime therefore follows the direction of
the proxy.

\clearpage

\clearpage
\section{Complete Evaluation Results}
\label{sec:app-full-results}

This section reports the numerical results underlying the main experiment
figures. Within each student scale, all preference-stage variants share the
prompts, chosen responses, SeqKD initialization, objective, optimizer budget,
and evaluation protocol; only the reject source changes.
\Cref{tab:kodcode_heldout_raw} reports KoDCode held-out pass@1, avg@4, and
pass@4 for every reject construction, and \Cref{tab:kodcode_external_pass1}
reports external pass@1 on BCB-C, BCB-I, HumanEval, and MBPP.
% \Cref{tab:in-domain-reliability-coverage-summary} summarizes the median effect
% of subscale Base sources on avg@4, which measures reliability over four draws,
% and on pass@4, which records whether a problem is solved at least once.
% \Cref{tab:kodcode-sample-gain-coverage} decomposes the pass@4 change of the
% selected Base source into problems newly solved and problems lost.
\Cref{tab:math500_aime_avg4} reports MATH-500 and AIME avg@4; among the 14
Qwen configurations with a smaller Base source, $13/14$ exceed
Continued-SFT on MATH-500 and $12/14$ on AIME.
\clearpage

% ---------------------------------------------------------------------
% Compact KoDCode held-out results. Greedy pass@1 is taken from
% analysis_out/kodcode_indomain_source.tex. avg@4 and pass@4 are derived from four stochastic
% samples per problem under eval_results/kodcode_heldout_stochastic_pass4/.
% Base checkpoints were evaluated in a follow-up sweep under the same
% stochastic prodtocol.
% ---------------------------------------------------------------------
\begin{table*}[!t]
  \centering
  \caption{KoDCode held-out results. pass@1 uses greedy decoding; avg@4 is the
  mean success rate over four stochastic samples, and pass@4 is the fraction
  of problems solved by at least one sample. Stochastic decoding uses
  temperature 0.6 and top-$p=1.0$. All values are percentages. Bold marks the
  best result within each student and metric.}
  \label{tab:kodcode_heldout_raw}

  \begin{tabular}{llcccccc}
    \toprule
    & & \multicolumn{4}{c}{pass@1} & \multicolumn{2}{c}{Stochastic} \\
    \cmidrule(lr){3-6} \cmidrule(lr){7-8}
    Method & Reject Source & Overall & Easy & Medium & Hard & avg@4 & pass@4 \\
    \midrule
    \multicolumn{8}{c}{\emph{Student: Qwen2.5-7B-Instruct}} \\
    \midrule
    Base & -- & 48.9 & 62.3 & 45.3 & 40.2 & 47.9 & 59.8 \\
    SeqKD & -- & 58.8 & 70.4 & 56.2 & 50.7 & 55.6 & 69.4 \\
    Continued-SFT & -- & 59.6 & 71.6 & 57.0 & 52.0 & 56.3 & 70.0 \\
    \cmidrule(lr){1-8}
    \multirow{8}{*}{DPO}
      & 0.5B & \textbf{60.5} & \textbf{72.0} & \textbf{57.5} & 53.1 & \textbf{58.5} & \textbf{70.5} \\
      & 1.5B & 59.8 & 70.8 & 57.0 & 52.7 & 57.4 & 70.2 \\
      & 3B   & 60.1 & 71.2 & 56.7 & \textbf{53.5} & 57.7 & 70.0 \\
      & 7B(\emph{Vanilla Self})   & 58.9 & 70.1 & 56.0 & 51.6 & 54.6 & 68.9 \\
      % & 14B  & 58.3 & 70.1 & 55.2 & 50.9 & 56.1 & 68.6 \\
      % & 32B  & 58.0 & 69.3 & 54.4 & 51.5 & 56.2 & 68.9 \\
      % & 72B  & 59.5 & 70.4 & 56.9 & 52.1 & 56.4 & 69.9 \\
      & 7B(\emph{Post-SeqKD Self}) & 58.6 & 69.5 & 56.0 & 50.7 & 52.7 & 69.9 \\

    \midrule
    \multicolumn{8}{c}{\emph{Student: Qwen2.5-14B-Instruct}} \\
    \midrule
    Base & -- & 50.9 & 63.3 & 47.4 & 43.1 & 50.4 & 63.4 \\
    SeqKD & -- & 61.7 & 72.6 & 58.8 & 54.7 & 60.2 & 72.4 \\
    Continued-SFT & -- & 62.5 & 73.7 & 59.6 & 55.3 & 60.9 & 72.7 \\
    \cmidrule(lr){1-8}
    \multirow{8}{*}{DPO}
      & 0.5B & 64.2 & \textbf{74.9} & 61.9 & 56.5 & 62.5 & 73.5 \\
      & 1.5B & 63.6 & 73.2 & 61.0 & 57.5 & 62.0 & 73.2 \\
      & 3B   & 64.3 & 74.8 & 61.6 & 57.6 & \textbf{62.6} & \textbf{74.0} \\
      & 7B   & \textbf{64.4} & 73.9 & \textbf{62.2} & \textbf{57.9} & 61.6 & 73.1 \\
      & 14B(\emph{Vanilla Self})  & 62.0 & 72.7 & 59.1 & 55.0 & 59.6 & 72.3 \\
      % & 32B  & 62.3 & 72.7 & 59.4 & 55.8 & 60.7 & 73.0 \\
      % & 72B  & 63.1 & 73.7 & 60.5 & 56.2 & 60.7 & 73.1 \\
      & 14B(\emph{Post-SeqKD Self}) & 63.3 & 73.5 & 60.4 & 57.1 & 58.2 & 73.7 \\

    \midrule
    \multicolumn{8}{c}{\emph{Student: Qwen2.5-32B-Instruct}} \\
    \midrule
    Base & -- & 57.3 & 67.9 & 53.9 & 51.3 & 56.5 & 67.6 \\
    SeqKD & -- & 64.1 & 74.5 & 61.2 & 57.4 & 62.9 & 73.7\\
    Continued-SFT & -- & 66.0 & 75.4 & 65.0 & 58.1 & 64.5 & 74.5 \\
    \cmidrule(lr){1-8}
    \multirow{8}{*}{DPO}
      & 0.5B & 65.8 & 75.2 & 62.5 & \textbf{60.8} & 65.1 & 74.8 \\
      & 1.5B & 66.0 & 73.6 & 62.8 & 60.7 & 64.7 & 74.4 \\
      & 3B   & 66.2 & 75.4 & 63.4 & \textbf{60.8} & \textbf{65.5} & 74.9 \\
      & 7B   & \textbf{66.9} & 75.9 & 64.8 & 60.5 & \textbf{65.5} & \textbf{76.1} \\
      & 14B  & 66.6 & \textbf{76.1} & 64.2 & 60.2 & 65.3 & 75.6 \\
      & 32B(\emph{Vanilla Self})  & 65.4 & 75.2 & 62.8 & 59.0 & 63.9 & 75.1 \\
      % & 72B  & 66.1 & 74.9 & 64.2 & 59.8 & 64.4 & 75.5 \\
      & 32B(\emph{Post-SeqKD Self}) & 65.8 & 75.0 & \textbf{65.3} & 58.4 & 63.1 & 76.0 \\

    \midrule
    \multicolumn{8}{c}{\emph{Student: Qwen2.5-72B-Instruct}} \\
    \midrule
    Vanilla & -- & 60.0 & 70.1 & 57.1 & 53.7 & 59.8 & 67.3 \\
    SeqKD & -- & 65.1 & 75.3 & 63.7 & 56.7 & 65.2 & 75.4 \\
    Continued-SFT & -- & 66.5 & 76.3 & 64.5 & 59.5 & 65.0 & 75.9 \\
    \cmidrule(lr){1-8}
    \multirow{8}{*}{DPO}
      & 0.5B & 68.3 & 77.8 & \textbf{66.1} & 61.7 & 67.3 & 76.6 \\
      & 1.5B & 68.5 & 76.9 & 66.0 & \textbf{63.6} & 67.3 & 76.3 \\
      & 3B   & \textbf{68.6} & \textbf{77.9} & 66.0 & 62.8 & 67.3 & 76.6 \\
      & 7B   & 68.3 & 77.8 & 65.8 & 62.2 & \textbf{67.5} & \textbf{77.2} \\
      & 14B  & 68.1 & 76.9 & 65.2 & 63.1 & 66.8 & 76.3 \\
      & 32B  & 68.2 & 77.1 & 65.5 & 62.9 & 66.7 & 76.6 \\
      & 72B(\emph{Vanilla Self})  & 68.3 & 77.3 & 65.9 & 62.3 & 65.9 & 76.6 \\
      & 72B(\emph{Post-SeqKD Self}) & 68.0 & 76.5 & \textbf{66.1} & 61.9 & 65.1 & 76.4 \\
    \bottomrule
  \end{tabular}
\end{table*}

\clearpage

% ---------------------------------------------------------------------
% External pass@1 results for the 36 final KoDCode preference-distillation
% endpoints. BCB-C/BCB-I values are the first-sample pass@1 statistics from
% the four-sample evaluations archived in
% analysis_out/kodcode_external_bcb_source.tex. HumanEval/MBPP
% values are greedy pass@1 from
% eval_results/bench_kodcode_preference_distillation_humaneval_mbpp/.
% Base checkpoints are omitted because matching MBPP evaluations are not
% available; every displayed row has results on all four benchmarks.
% ---------------------------------------------------------------------
\begin{table*}[!t]
  \centering
  \caption{External pass@1 across four code-generation benchmarks. Weighted Avg. weights each benchmark by its
  number of problems: 1{,}140 each for BCB-C and BCB-I, 164 for HumanEval,
  and 257 for MBPP. All values are percentages. Bold marks the best result
  within each student and column.}
  \label{tab:kodcode_external_pass1}

  \small
  \begin{tabular}{llccccc}
    \toprule
    Method & Reject Source & BCB-C & BCB-I & HumanEval & MBPP & Weighted Avg. \\
    \midrule
    \multicolumn{7}{c}{\emph{Student: Qwen2.5-7B-Instruct}} \\
    \midrule
    Base & -- & 42.3 & 29.1 & 84.1 & 79.0 & 42.8 \\
    SeqKD & -- & 43.4 & 32.5 & 83.3 & 79.1 & 44.6 \\
    Continued-SFT & -- & 43.9 & 33.6 & 82.3 & 79.1 & 45.2 \\
    \cmidrule(lr){1-7}
    \multirow{5}{*}{DPO}
      & 0.5B & \textbf{45.2} & 35.4 & \textbf{84.8} & 78.2 & \textbf{46.6} \\
      & 1.5B & 45.0 & 33.3 & 82.3 & \textbf{80.2} & 45.7 \\
      & 3B   & 44.8 & 34.0 & 80.5 & 77.4 & 45.5 \\
      & 7B(\emph{Vanilla Self})   & 42.6 & 33.2 & 79.9 & 76.7 & 44.1 \\
      % & 14B  & 45.0 & \textbf{35.9} & 76.8 & 77.8 & 46.2 \\
      % & 32B  & 42.8 & 33.3 & 78.0 & 74.7 & 44.0 \\
      % & 72B  & 44.8 & 31.5 & 80.5 & 75.9 & 44.3 \\
      & 7B((\emph{Post-SeqKD Self})) & 40.9 & 30.1 & 80.5 & 79.4 & 42.4 \\

    \midrule
    \multicolumn{7}{c}{\emph{Student: Qwen2.5-14B-Instruct}} \\
    \midrule
    Base & -- & 48.2 & 35.5 & 82.9 & 79.0 & 47.9 \\
    SeqKD & -- & 49.1 & 37.3 & 82.8 & 81.4 & 49.2 \\
    Continued-SFT & -- & 49.4 & 39.2 & 81.7 & 82.5 & 50.2 \\
    \cmidrule(lr){1-7}
    \multirow{6}{*}{DPO}
      & 0.5B & 50.0 & \textbf{40.4} & 82.3 & 82.5 & 51.0 \\
      & 1.5B & 49.3 & 40.2 & \textbf{84.1} & \textbf{83.3} & 50.8 \\
      & 3B   & \textbf{52.4} & 40.3 & 83.5 & 82.5 & \textbf{52.0} \\
      & 7B   & 48.9 & 39.2 & 81.7 & 81.3 & 49.9 \\
      & 14B(\emph{Vanilla Self})  & 48.4 & 37.4 & 80.5 & 80.2 & 48.7 \\
      % & 32B  & 47.4 & 39.6 & 81.7 & 82.5 & 49.5 \\
      % & 72B  & 50.6 & 38.9 & 81.1 & 80.5 & 50.4 \\
      & 14B(\emph{Post-SeqKD Self}) & 46.0 & 37.2 & 86.0 & 81.3 & 48.1 \\

    \midrule
    \multicolumn{7}{c}{\emph{Student: Qwen2.5-32B-Instruct}} \\
    \midrule
    Base & -- & 52.8 & 38.6 & 89.6 & \textbf{87.2} & 52.3 \\
    SeqKD & -- & 53.2 & 39.9 & 86.7 & 86.8 & 52.8 \\
    Continued-SFT & -- & 53.0 & 41.8 & 87.2 & 86.4 & 53.5 \\
    \cmidrule(lr){1-7}
    \multirow{7}{*}{DPO}
      & 0.5B & 52.7 & 40.6 & 87.8 & 85.3 & 52.8 \\
      & 1.5B & 53.3 & 41.7 & 86.0 & 85.9 & 53.5 \\
      & 3B   & 54.9 & 42.1 & \textbf{90.9} & 85.3 & 54.6 \\
      & 7B   & \textbf{56.3} & 42.8 & 87.2 & 86.8 & \textbf{55.4} \\
      & 14B  & 53.2 & \textbf{44.0} & 86.6 & 85.3 & 54.4 \\
      & 32B(\emph{Vanilla Self})  & 54.4 & 41.1 & 87.8 & 86.4 & 53.9 \\
      % & 72B  & 54.0 & 41.8 & 86.0 & 86.4 & 53.9 \\
      & 32B(\emph{Post-SeqKD Self}) & 51.0 & 41.4 & 88.4 & 84.8 & 52.4 \\

    \midrule
    \multicolumn{7}{c}{\emph{Student: Qwen2.5-72B-Instruct}} \\
    \midrule
    Base & -- & 54.2 & 43.4 & 87.8 & 89.2 & 54.8 \\
    SeqKD & -- & 55.1 & 43.7 & 86.5 & 87.4 & 55.3 \\
    Continued-SFT & -- & 55.9 & \textbf{44.3} & 86.0 & 86.4 & 55.7 \\
    \cmidrule(lr){1-7}
    \multirow{8}{*}{DPO}
      & 0.5B & 55.8 & 43.7 & 87.8 & 88.6 & 55.8 \\
      & 1.5B & 56.1 & 43.2 & \textbf{90.2} & 89.4 & 55.9 \\
      & 3B   & 55.7 & 42.5 & 89.6 & \textbf{89.5} & 55.4 \\
      & 7B   & 56.8 & 43.9 & 87.8 & 89.0 & 56.3 \\
      & 14B  & \textbf{57.6} & 43.9 & 89.0 & 89.0 & \textbf{56.7} \\
      & 32B  & 56.6 & 42.9 & 89.6 & 88.2 & 55.8 \\
      & 72B(\emph{Vanilla Self})  & 56.3 & 40.8 & 86.0 & 87.4 & 54.5 \\
      & 72B(\emph{Post-SeqKD Self}) & 54.5 & 42.4 & 88.4 & 86.8 & 54.5 \\
    \bottomrule
  \end{tabular}
\end{table*}
\clearpage
% \input{sections/tables/indomain_reliability_coverage_summary}

% \input{sections/tables/kodcode_sample_gain_coverage}

% \begin{figure*}[t]
%   \centering
%   \includegraphics[width=0.8\textwidth]{sections/figures/reject_direct_effects.pdf}
%   \caption{Repeat success and observed coverage across reject constructions.
%   Strictly smaller Base rejects uniformly separate from Self on avg@4, while
%   pass@4 does not exhibit the same class ordering.}
%   \label{fig:app-reject-direct-effects}
% \end{figure*}

% MATH-500 and AIME avg@4 results table.
% Auto-maintained from eval_results/openr1_aime_avgk4_{7b,14b,32b,72b}/bench_*_{math500,aime}.json
% (avg@k harness: src/eval_bench/run_bench.py, K=4, temperature 0.6, thinking OFF).
% Bold = best (highest avg@4) within each student block for that benchmark column.
% Blank = not yet run.
\begin{table*}[t]
\centering
\caption{MATH-500 and AIME avg@4 over four samples.}
\label{tab:math500_aime_avg4}
\small
\begin{tabular}{llcc}
\toprule
Method & Reject Source & MATH-500 avg@4 & AIME avg@4 \\
\midrule
Teacher (DeepSeek-R1) & -- & 97.3 & 74.9 \\
\midrule
\multicolumn{4}{c}{\emph{Student: Qwen2.5-72B-Instruct}} \\
\midrule
Base   & --   & 84.0 & 15.8 \\
SeqKD  & --   & 91.0 & 37.9 \\
Continued-SFT & --   &  91.2 &  37.5 \\
\midrule
\multirow{9}{*}{DPO}
       & 0.5B     & 91.6 & 40.0 \\
       & 7B       & 91.5 & 37.1 \\
       & 14B      & 91.7 & \textbf{40.4} \\
       & 32B      & \textbf{92.7} & 40.4 \\
       & 72B(\emph{Vanilla Self}) & 91.8 & 35.4 \\
       & 72B(\emph{Post-SeqKD Self}) & 91.8 & 37.5 \\
       & Llama-1B & 91.7 & 34.2 \\
       & Llama-3B & 92.0 & 39.6 \\
       & Llama-8B & 92.0 & 35.4 \\
%        & Llama-70B & 91.6 & 36.7 \\
\midrule
\multicolumn{4}{c}{\emph{Student: Qwen2.5-32B-Instruct}} \\
\midrule
Base   & --   & 82.5 & 13.8 \\
SeqKD  & --   & 87.8 & 29.2 \\
Continued-SFT & --   & 87.6 & 29.2 \\
\midrule
\multirow{10}{*}{DPO}
       & 0.5B     & 88.4 & 32.1 \\
       & 7B       & 88.4 & \textbf{35.4} \\
       & 14B      & 89.1 & 32.1 \\
       & 32B(\emph{Vanilla Self})      & 88.0 & 30.8 \\
       & 32B(\emph{Post-SeqKD Self})      & 87.9 & 30.4 \\
       % & 72B      & 88.8 & 30.8 \\
       & Llama-1B & 88.6 & 29.2 \\
       & Llama-3B & \textbf{89.5} & 32.5 \\
       & Llama-8B & 89.1 & 32.5 \\
       % & Llama-70B & 89.4 & 30.4 \\
\midrule
\multicolumn{4}{c}{\emph{Student: Qwen2.5-14B-Instruct}} \\
\midrule
Base   & --   & 78.7 & 13.3 \\
SeqKD  & --   & 81.5 & 18.8 \\
Continued-SFT & --   & 82.1 & 19.6 \\
\midrule
\multirow{10}{*}{DPO}
       & 0.5B     & 83.1 & \textbf{22.9} \\
       & 1.5B     & 82.9 & \textbf{22.9} \\
       & 3B       & 83.2 & 20.0 \\
       & 7B       & 82.0 & \textbf{22.9} \\
       & 14B(\emph{Vanilla Self})      & 81.3 & 20.0 \\
       & 14B(\emph{Post-SeqKD Self})      & 80.9 & 19.6 \\
       % & 72B      & 82.3 & 22.5 \\
       & Llama-1B & 83.9 & 20.8 \\
       & Llama-3B & \textbf{84.6} & 20.4 \\
       & Llama-8B & 83.9 & 21.3 \\
       % & Llama-70B & 82.5 & 22.5 \\
\midrule
\multicolumn{4}{c}{\emph{Student: Qwen2.5-7B-Instruct}} \\
\midrule
Base       & --   & \textbf{76.0} & 7.9 \\
SeqKD  & --   & 78.5 & 10.8 \\
Continued-SFT & --   & 79.2 & 11.7 \\
\midrule
\multirow{5}{*}{DPO}
       & 0.5B     & 79.9 & 16.7 \\
       & 1.5B     & 80.2 & \textbf{22.9} \\
       & 3B       & 79.6 & 11.7 \\
       & 7B(\emph{Vanilla Self}) & 78.7 & 14.2 \\
       & 7B(\emph{Post-SeqKD Self}) & 78.2 & 12.9\\
% \midrule
% \multirow{8}{*}{\shortstack[l]{DPO \\ (no SFT, from Base)}}
%        & 0.5B     & 77.5 & \textbf{14.6} \\
%        & 7B       & 79.6 & 12.9 \\
%        & 14B      & 78.5 & 10.8 \\
%        & 32B      & \textbf{80.0} & 14.2 \\
%        & 72B      & 75.5 & 10.4 \\
%        & Llama-1B & 78.3 & 10.4 \\
%        & Llama-8B & 77.7 & 14.6 \\
%        & Llama-70B & 78.0 & 11.7 \\
% \midrule
% \multirow{7}{*}{\shortstack[l]{re-SODA \\ (LD-DPO)}}
%        & 0.5B     & \textbf{54.6} & 15.0 \\
%        & 7B       & 49.7 & 16.7 \\
%        & 14B      & 44.5 & 11.7 \\
%        & 32B      & 50.3 & 16.7 \\
%        & 72B      &  & \textbf{17.5} \\
%        & Llama-1B &  & 0.0 \\
%        & Llama-8B &  & 5.8 \\
\bottomrule
\end{tabular}
\end{table*}
\clearpage

\subsection{Cross-Family Reject Sources}
\label{sec:app-cross-family}

\begin{figure*}[t]
  \centering
  \includegraphics[width=\textwidth]{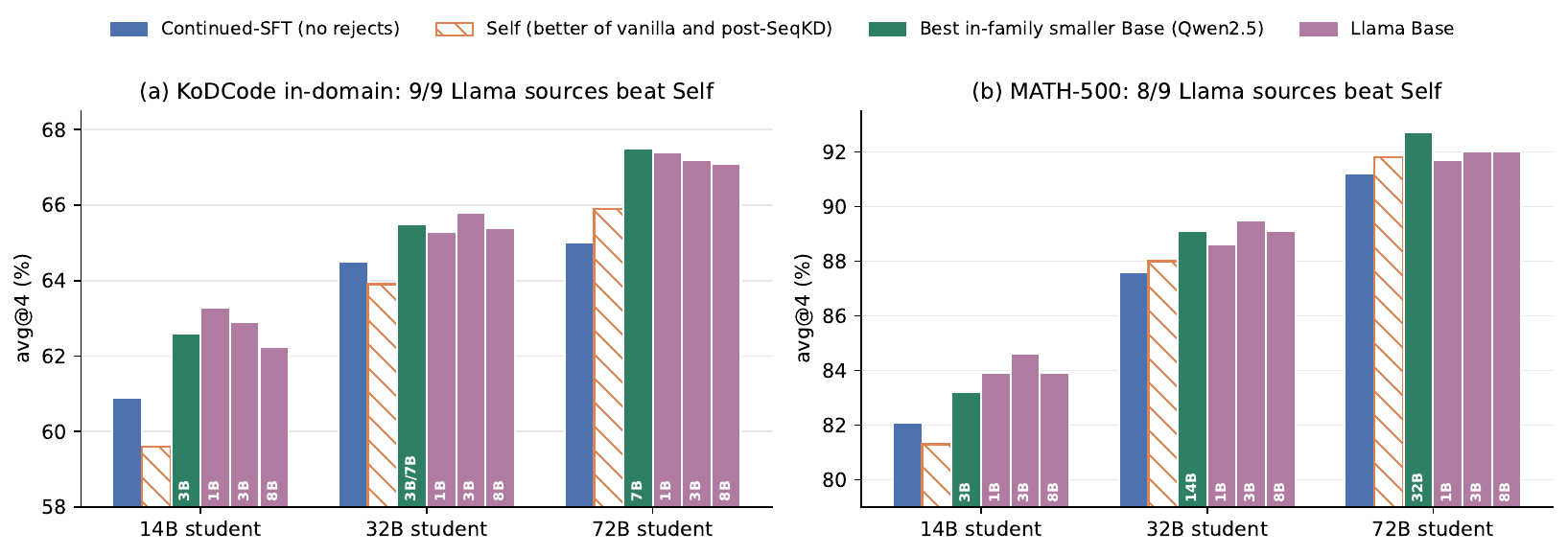}
  \caption{Cross-family reject sources. Bars report avg@4 on (a) KoDCode
  in-domain and (b) MATH-500 for the 14B, 32B, and 72B Qwen2.5 students. Self
  is the better of vanilla Self and post-SeqKD Self. Labels on the green bars
  identify the best Qwen2.5 smaller Base source; labels on the purple bars
  identify the Llama source.}
  \label{fig:app-cross-family}
\end{figure*}

To test whether the smaller Base advantage depends on generating rejects
within the student's model family, we replace the Qwen2.5 reject source with
Llama-3.2-1B-Instruct, Llama-3.2-3B-Instruct, or Llama-3.1-8B-Instruct for
the 14B, 32B, and 72B Qwen2.5 students. All three are smaller Base sources
for these students. \Cref{fig:app-cross-family} compares them with
Continued-SFT, the better of the two Self constructions, and the best
Qwen2.5 smaller Base source at the same student scale.

Llama rejects outperform Self in all nine code configurations and in eight
of nine MATH-500 configurations; the exception is Llama-1B for the 72B
student on MATH-500 ($91.7\%$ versus $91.8\%$).

\clearpage

\newpage
\section{Construction Intervention Details}
\label{sec:app-interventions}

This section reports the complete results of the construction interventions
in \Cref{sec:theory-interventions}: randomized source assignment (P1), prompt
reassignment and lexical permutation (P2), and reference likelihood
reselection (P3), together with the natural source coupling analysis that
motivates P3. Within each intervention, all arms share the student, prompts,
chosen responses, SeqKD initialization, optimizer schedule, and evaluation
protocol; only the reject construction changes.

\subsection{Randomized Source Assignment}
\label{sec:app-mixture-results}

We implement the randomized source assignment operator in
\Cref{eq:theory-mixture} with three interventions
(\Cref{tab:randomized-source-assignment}). For the 14B student, rejects come
from vanilla Self or from either Llama-3.2-1B-Instruct or
Qwen2.5-3B-Instruct; for the 72B student, they come from
vanilla Self or Qwen2.5-7B-Instruct. In every intervention, avg@4 rises with the Base share: from $59.730$ to
$63.280$ with Llama-3.2-1B-Instruct, from $59.575$ to $62.630$ with
Qwen2.5-3B-Instruct, and from $65.850$ to $67.530$ with Qwen2.5-7B-Instruct,
passing through $66.685$ at $50\%$. Source composition therefore controls
the attainable utility continuously across model families and student
scales.

\begin{table*}[t]
  \centering
  \caption{Complete randomized source assignment results. The Base share is
  the proportion of prompts assigned rejects from the indicated smaller Base source. All remaining prompts use vanilla Self rejects from the
  student-scale model.}
  \label{tab:randomized-source-assignment}
  \small
  \begin{tabular}{llrrr}
    \toprule
    Student & Smaller Base source & Base share & Code avg@4 & Code pass@4 \\
    \midrule
    \multirow{5}{*}{14B} & \multirow{5}{*}{Llama-3.2-1B-Instruct}
      & $0\%$   & 59.730 & 72.380 \\
    & & $25\%$  & 61.575 & 73.000 \\
    & & $50\%$  & 62.035 & 72.760 \\
    & & $75\%$  & 62.670 & 73.380 \\
    & & $100\%$ & 63.280 & 73.180 \\
    \midrule
    \multirow{5}{*}{14B} & \multirow{5}{*}{Qwen2.5-3B-Instruct}
      & $0\%$   & 59.575 & 72.280 \\
    & & $25\%$  & 60.865 & 73.100 \\
    & & $50\%$  & 61.800 & 73.860 \\
    & & $75\%$  & 62.075 & 73.300 \\
    & & $100\%$ & 62.630 & 73.960 \\
    \midrule
    \multirow{3}{*}{72B} & \multirow{3}{*}{Qwen2.5-7B-Instruct}
      & $0\%$   & 65.850 & 76.560 \\
    & & $50\%$  & 66.685 & 76.580 \\
    & & $100\%$ & 67.530 & 77.240 \\
    \bottomrule
  \end{tabular}
\end{table*}

\subsection{Prompt Reassignment and Lexical Permutation}
\label{sec:app-localization-details}
\label{sec:app-structure-details}

\begin{figure*}[t]
  \centering
  \includegraphics[width=\textwidth]{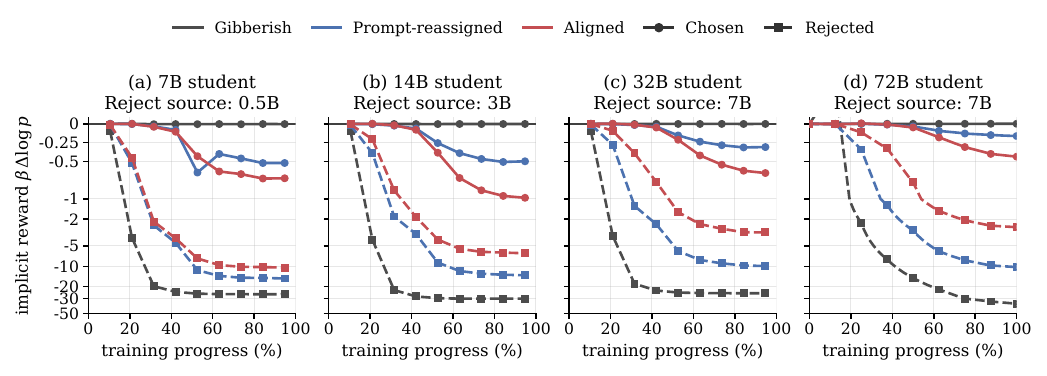}
  \caption{Implicit reward dynamics for gibberish, prompt reassigned, and
  prompt-matched Base rejects at each student scale. Gibberish is suppressed most
  strongly but does not recover the utility of real Base rejects.}
  \label{fig:app-control-reward-dynamics}
\end{figure*}

Prompt reassignment assigns each real Base reject to a different
prompt. To avoid trivial interface mismatches, we rename each
reassigned reject's entry-point function to match the new prompt,
and we retain only rejects that fail at least one of the new
prompt's unit tests, so every reassigned reject remains verified
incorrect. \Cref{tab:kodcode-reject-control-ladder} reports the Continued-SFT, Gibberish,
prompt-reassigned, and aligned endpoints at every student scale, and
\Cref{fig:app-control-reward-dynamics} shows the corresponding implicit reward
trajectories.

% Cross-scale controls that localize the stable strict-subscale effect.
\begin{table*}[t]
  \centering
  \caption{KoDCode avg@4 across the reject-control ladder. Prompt-reassigned
  arms use real Base rejects reassigned across prompts. Each row is one
  completed prompt-reassignment experiment and reports its Continued-SFT,
  gibberish, reassigned, and aligned settings.}
  \label{tab:kodcode-reject-control-ladder}
  \small
  \begin{tabular}{llrrrr}
    \toprule
    Student & Reject source & Continued-SFT & Gibberish
      & Prompt-reassigned & Aligned \\
    \midrule
    7B  & Qwen2.5-0.5B-Instruct & 56.30 & 57.37 & 58.65 & 58.53 \\
    14B & Qwen2.5-3B-Instruct   & 60.89 & 60.92 & 61.97 & 62.63 \\
    32B & Qwen2.5-7B-Instruct   & 64.50 & 64.52 & 65.25 & 65.53 \\
    72B & Qwen2.5-7B-Instruct   & 65.00 & 65.24 & 66.36 & 67.53 \\
    \bottomrule
  \end{tabular}
\end{table*}

At every student scale, lexical permutation starts from the corresponding
prompt-reassigned Base rejects. It preserves the generated code-token
inventory while destroying token order, Python syntax, coherent program
semantics, and prompt correspondence (\Cref{tab:structure-ablation}).
Lexical permutation improves avg@4 over Gibberish at all four scales and
recovers between $40.0\%$ and $92.5\%$ of the real prompt-reassigned gain. This
identifies code-domain lexical structure as a cross-scale component of Base
reject utility. At 14B, an additional AST scrambling intervention reaches
$61.840$ avg@4. It preserves Python syntax and common AST constructs while
disrupting program semantics.

\begin{table*}[t]
  \centering
  \vspace{-0.2cm}
  \caption{Cross-scale lexical-structure interventions. Recovery is the
  Lexical gain over Gibberish divided by the real prompt-reassigned gain over
  Gibberish.}
  \label{tab:structure-ablation}
  \small
  \begin{tabular}{lrrrrr}
    \toprule
    Student & Gibberish & Lexical & Real reassigned
      & Lexical $-$ Gibberish & Recovery \\
    \midrule
    7B  & 57.370 & 57.900 & 58.645
      & $+0.530$ & $41.6\%$ \\
    14B & 60.920 & 61.650 & 61.970
      & $+0.730$ & $69.5\%$ \\
    32B & 64.520 & 65.195 & 65.250
      & $+0.675$ & $92.5\%$ \\
    72B & 65.235 & 65.685 & 66.360
      & $+0.450$ & $40.0\%$ \\
    \bottomrule
  \end{tabular}
  \vspace{-0.4cm}
\end{table*}

\subsection{Reference Coupling of Natural Sources}
\label{sec:app-natural-coupling}

\begin{table*}[p]
  \centering
  \caption{Natural reject-source measurements underlying
  \Cref{fig:reference-coupling-evidence}a. Each source is scored on the same
  24,248 preference prompts under the corresponding student's SeqKD
  reference. Mean log likelihood is length normalized. Prompt $z$ is the mean
  reference likelihood after standardization within each prompt. Utility is
  KoDCode avg@4 on 5,000 held-out problems.}
  \label{tab:reference-coupling-natural-sources}
  \scriptsize
  \begin{tabular}{lllrrrr}
    \toprule
    Student & Reject source & Source class
      & Mean log $S$/token & Prompt $z$ & Mean tokens & avg@4 \\
    \midrule
    7B & Qwen2.5-0.5B-Instruct & Subscale Base
      & $-0.3725$ & $-1.048$ & 466.7 & 58.525 \\
    7B & Qwen2.5-1.5B-Instruct & Subscale Base
      & $-0.2527$ & $-0.113$ & 262.5 & 57.395 \\
    7B & Qwen2.5-3B-Instruct & Subscale Base
      & $-0.2851$ & $-0.435$ & 291.8 & 57.705 \\
    7B & Qwen2.5-7B-Instruct & Vanilla Self
      & $-0.1790$ & $+0.418$ & 284.9 & 54.550 \\
    7B & post-SeqKD 7B & Post-SeqKD Self
      & $-0.0915$ & $+1.179$ & 243.7 & 52.745 \\
    \midrule
    14B & Qwen2.5-0.5B-Instruct & Subscale Base
      & $-0.3848$ & $-1.074$ & 466.7 & 62.525 \\
    14B & Qwen2.5-1.5B-Instruct & Subscale Base
      & $-0.2680$ & $-0.140$ & 262.5 & 61.995 \\
    14B & Qwen2.5-3B-Instruct & Subscale Base
      & $-0.2922$ & $-0.398$ & 291.8 & 62.630 \\
    14B & Qwen2.5-7B-Instruct & Subscale Base
      & $-0.2206$ & $+0.179$ & 284.9 & 61.645 \\
    14B & Qwen2.5-14B-Instruct & Vanilla Self
      & $-0.2255$ & $+0.127$ & 304.7 & 59.575 \\
    14B & post-SeqKD 14B & Post-SeqKD Self
      & $-0.0874$ & $+1.306$ & 240.9 & 58.155 \\
    \midrule
    32B & Qwen2.5-0.5B-Instruct & Subscale Base
      & $-0.3957$ & $-1.130$ & 466.7 & 65.135 \\
    32B & Qwen2.5-1.5B-Instruct & Subscale Base
      & $-0.2784$ & $-0.203$ & 262.5 & 64.700 \\
    32B & Qwen2.5-3B-Instruct & Subscale Base
      & $-0.3045$ & $-0.467$ & 291.8 & 65.515 \\
    32B & Qwen2.5-7B-Instruct & Subscale Base
      & $-0.2234$ & $+0.169$ & 284.9 & 65.530 \\
    32B & Qwen2.5-14B-Instruct & Subscale Base
      & $-0.2570$ & $-0.102$ & 304.7 & 65.285 \\
    32B & Qwen2.5-32B-Instruct & Vanilla Self
      & $-0.1921$ & $+0.422$ & 266.4 & 63.880 \\
    32B & post-SeqKD 32B & Post-SeqKD Self
      & $-0.0834$ & $+1.312$ & 242.0 & 63.070 \\
    \midrule
    72B & Qwen2.5-0.5B-Instruct & Subscale Base
      & $-0.4034$ & $-1.204$ & 466.7 & 67.290 \\
    72B & Qwen2.5-1.5B-Instruct & Subscale Base
      & $-0.2840$ & $-0.273$ & 262.5 & 67.300 \\
    72B & Qwen2.5-3B-Instruct & Subscale Base
      & $-0.3130$ & $-0.553$ & 291.8 & 67.270 \\
    72B & Qwen2.5-7B-Instruct & Subscale Base
      & $-0.2290$ & $+0.096$ & 284.9 & 67.530 \\
    72B & Qwen2.5-14B-Instruct & Subscale Base
      & $-0.2685$ & $-0.213$ & 304.7 & 66.800 \\
    72B & Qwen2.5-32B-Instruct & Subscale Base
      & $-0.2262$ & $+0.124$ & 266.4 & 66.670 \\
    72B & Qwen2.5-72B-Instruct & Vanilla Self
      & $-0.1452$ & $+0.754$ & 239.6 & 65.850 \\
    72B & post-SeqKD 72B & Post-SeqKD Self
      & $-0.0821$ & $+1.269$ & 239.3 & 65.070 \\
    \bottomrule
  \end{tabular}
\end{table*}

\Cref{tab:reference-coupling-natural-sources} reports every point used in the
natural-source analysis. For each student scale, the likelihood measurements
use the same 24,248 preference prompts and the exact SeqKD model used as the
DPO reference. Downstream utility uses the same 5,000 held-out problems and
four samples per problem. The negative association is preserved under prompt standardization and every
leave-one-source-out analysis (\Cref{tab:reference-coupling-correlations}). 
% Removing post-SeqKD Self weakens the
% association, especially at 32B, while retaining a negative point estimate at
% all scales. 
Reference likelihood therefore organizes the Base-versus-Self boundary. 
% The smaller Base only analysis is not uniformly negative, so it does not
% provide a complete ranking rule within the smaller Base class.

\begin{table*}[t]
  \centering
  \caption{Cross-source association between mean reference likelihood and
  downstream avg@4. Exact $p$-values enumerate all assignments of the
  observed utilities to sources; $p_{-}$ is the one-sided test for a negative
  association. Prompt $z$ uses within-prompt standardized reference
  likelihood. Leave-one-out (LOO)
  reports the range of Pearson correlations after removing each source in
  turn.}
  \label{tab:reference-coupling-correlations}
  \small
  \begin{tabular}{lrrrrrrr}
    \toprule
    Student & $n$ & $r$ & $\rho$ & $p_{-}$ & $p_{2}$ & Prompt-$z$ $r$
      & LOO $r$ range \\
    \midrule
    7B  & 5 & $-0.967$ & $-1.000$ & $0.0083$ & $0.0083$ & $-0.968$
      & $[-0.988,-0.920]$ \\
    14B & 6 & $-0.868$ & $-0.886$ & $0.0083$ & $0.0083$ & $-0.871$
      & $[-0.909,-0.645]$ \\
    32B & 7 & $-0.752$ & $-0.464$ & $0.0298$ & $0.0484$ & $-0.770$
      & $[-0.863,-0.405]$ \\
    72B & 8 & $-0.836$ & $-0.690$ & $0.0057$ & $0.0076$ & $-0.842$
      & $[-0.931,-0.684]$ \\
    \bottomrule
  \end{tabular}
\end{table*}

\subsection{Reference Likelihood Reselection}
\label{sec:app-coupling-reselection}

For each source, we sample eight candidates for each preference prompt and
score their length-normalized likelihood under the 14B SeqKD reference within
the prompt-specific bank. Shared Gumbel draws construct reference-repelled,
native, and reference-attracted datasets from each fixed bank. A prompt is then kept only if all 3 selected rejects run to completion, pass exactly the same fraction of unit tests, and differ in length by at most 32 tokens. We retain the intersection of prompts satisfying these criteria across all reject sources, yielding a shared set of 20,136 prompts used for every source and all three selection rules. Repelled selection improves on Native for all five sources (\Cref{tab:reference-coupling-reselection}). Attracted selection
reduces utility for all four smaller Base sources. Combining the 14B repel
effect with the attracted effect from each smaller source yields positive
double-dissociation interactions of $+0.815$, $+1.000$, $+0.695$, and
$+1.090$ points for 0.5B, 1.5B, 3B, and 7B sources. Across the 15 arms, after
centering coupling and utility within each generator, their Pearson and
Spearman associations are $0.868$ and $0.936$, respectively.

\begin{table*}[t]
  \centering
  \caption{Reference-likelihood reselection for the Qwen2.5-14B-Instruct
  student. Coupling entries are the negative mean length-normalized log
  likelihood under the 14B SeqKD reference, so larger values indicate lower
  reference coupling. Utility entries are KoDCode avg@4; the final three
  columns give the within-source contrasts. Native arms are trained on the
  reselection candidate banks, so their values differ from the natural source
  endpoints in \Cref{tab:reference-coupling-natural-sources}.}
  \label{tab:reference-coupling-reselection}
  \scriptsize
  \setlength{\tabcolsep}{4pt}
  \begin{tabular}{lrrrrrrrrr}
    \toprule
    & \multicolumn{3}{c}{Coupling ($-\log S$/token)}
      & \multicolumn{3}{c}{avg@4}
      & \multicolumn{3}{c}{Contrast} \\
    \cmidrule(lr){2-4}\cmidrule(lr){5-7}\cmidrule(lr){8-10}
    Reject source & Repelled & Native & Attracted
      & Repelled & Native & Attracted
      & R $-$ N & A $-$ N & R $-$ A \\
    \midrule
    Qwen2.5-0.5B-Instruct & 0.690 & 0.519 & 0.359
      & 63.335 & 63.075 & 62.750 & $+0.260$ & $-0.325$ & $+0.585$ \\
    Qwen2.5-1.5B-Instruct & 0.487 & 0.344 & 0.238
      & 63.135 & 62.390 & 61.880 & $+0.745$ & $-0.510$ & $+1.255$ \\
    Qwen2.5-3B-Instruct   & 0.492 & 0.363 & 0.260
      & 63.120 & 62.860 & 62.655 & $+0.260$ & $-0.205$ & $+0.465$ \\
    Qwen2.5-7B-Instruct   & 0.320 & 0.243 & 0.183
      & 62.100 & 62.070 & 61.470 & $+0.030$ & $-0.600$ & $+0.630$ \\
    Qwen2.5-14B-Instruct  & 0.363 & 0.263 & 0.191
      & 59.905 & 59.415 & 59.455 & $+0.490$ & $+0.040$ & $+0.450$ \\
    \bottomrule
  \end{tabular}
\end{table*}

\clearpage
\section{Additional Explanation Boundaries}
\label{sec:app-explanation-boundaries}

The direct construction interventions provide the main evidence.
This section reports complementary diagnostics that test whether
optimization summaries recover the same source ordering.

\subsection{DPO Diagnostic Definitions}
\label{sec:app-dpo-diagnostics}

We define the DPO diagnostics used in the fixed-pair and RC-DPO
analyses. DPO induces the implicit reward~\citep{dpo,gradientEntanglement}
\begin{equation}
  \widehat r_{\theta}(x,y)
  = \beta \log \frac{\pi_{\theta}(y\mid x)}
                          {\pi_{\mathrm{ref}}(y\mid x)}.
  \label{eq:dpo-implicit-reward}
\end{equation}
We write $\widehat r^{+}_{\theta}$ and $\widehat r^{-}_{\theta}$ for the
chosen and rejected implicit rewards. At DPO initialization,
$\pi_{\theta_0}=\pi_{\mathrm{ref}}$, so every pair has zero implicit margin
and logistic weight $1/2$. Define the mean chosen field and the mean rejected
field induced by source $q$ as
\begin{equation}
  F^+=\frac{1}{n}\sum_{i=1}^n
  \nabla_\theta\log\pi_{\theta_0}(y_i^+\mid x_i),
  \qquad
  F_q^-=\frac{1}{n}\sum_{i=1}^n
  \nabla_\theta\log\pi_{\theta_0}(y_{i,q}^-\mid x_i).
  \label{eq:chosen-reject-fields}
\end{equation}
The initialization gradient is
\begin{equation}
  \nabla_\theta\mathcal L_{\mathrm{DPO}}(\theta_0)
  =-\frac{\beta}{2}\left(F^+-F_q^-\right).
  \label{eq:dpo-initial-gradient}
\end{equation}
For exact random sampling from the reference, the score-function identity gives
\begin{equation}
  \E_{y\sim\pi_{\mathrm{ref}}(\cdot\mid x)}
  \!\left[\nabla_\theta\log\pi_{\theta_0}(y\mid x)\right]=0.
  \label{eq:score-function-reference}
\end{equation}

For the empirical field measurement, let $L_{i,q}^{-}$ denote the number of rejected-response tokens, and let $P$ denote a fixed CountSketch projection of the score gradient with respect to the final-layer parameters. We report the per-token rejected-score
field
\begin{equation}
  \widetilde F_q^-
  =\frac{1}{n}\sum_{i=1}^n
  P\!\left(
    \frac{1}{L_{i,q}^-}
    \nabla_\theta\log\pi_{\theta_0}(y_{i,q}^-\mid x_i)
  \right),
  \qquad
  N_q^-=\lVert\widetilde F_q^-\rVert_2.
  \label{eq:projected-reject-field}
\end{equation}
Under gradient descent, $-F_q^-$ is the first-step component that lowers
rejected likelihood. The projection norm is used only for within-student
comparisons under the same tangent space.

For a fixed chosen or rejected response, define the sequence log-likelihood
displacement at checkpoint $t$ as
\begin{equation}
  \Delta\ell_{i,t}^{\pm}
  =\log\pi_{\theta_t}(y_i^{\pm}\mid x_i)
  -\log\pi_{\theta_0}(y_i^{\pm}\mid x_i).
  \label{eq:fixed-pair-displacement}
\end{equation}
Following RC-DPO~\citep{rc-dpo}, Pathway III contains pairs for which
$\Delta\ell_{i,t}^{+}\geq0$ and $\Delta\ell_{i,t}^{-}\leq0$.

RC-DPO rescales the backward incentives that produce these dynamics. For
batch-mean sequence log probabilities $\ell_t^\pm$, standard DPO uses
$z_t=(\ell_t^+-\ell_{\mathrm{ref}}^+)
-(\ell_t^--\ell_{\mathrm{ref}}^-)$ and
$\mathcal L_t=-\log\sigma(\beta z_t)$. Let $c_t$ and $d_t$ be
log-EMA-smoothed norms of $\nabla\ell_t^+$ and $\nabla\ell_t^-$ on a
trainable output-head subspace, and set $\alpha_t=\sqrt{d_t/c_t}$. RC-DPO uses
\begin{equation}
  \widetilde\ell_t^+
  =\alpha_t\ell_t^+ +(1-\alpha_t)\operatorname{sg}(\ell_t^+),
  \qquad
  \widetilde\ell_t^-
  =\alpha_t^{-1}\ell_t^- +(1-\alpha_t^{-1})
    \operatorname{sg}(\ell_t^-).
  \label{eq:rc-dpo-rescaling}
\end{equation}
The forward log probabilities are unchanged, while the backward pass scales the
chosen and rejected gradients by $\alpha_t$ and $\alpha_t^{-1}$, respectively.

\subsection{Reject-Score Fields}

The student-local score-field analysis uses 128 shared prompts and two
independent CountSketch projections. Sequence-sum and per-token norms are
reported separately, and all comparisons are made within a student and
normalization view. The complete 14B and 72B measurements appear in
\Cref{tab:reject-score-field-summary}.

\begin{table*}[t]
  \centering
  \caption{Student-local reject-score fields at the shared preference-
  optimization initialization. Norms are averaged over two independent
  sketches. Differences are source minus post-SeqKD Self. Per-token norms and differences are scaled by
  $10^3$.}
  \label{tab:reject-score-field-summary}
  \scriptsize
  \setlength{\tabcolsep}{4pt}
  \begin{tabular}{lrrrrr}
    \toprule
    Reject source & Mean-field norm & $\Delta$ field &
    Coherence & Per-token norm & $\Delta$ per-token \\
    \midrule
    \multicolumn{6}{c}{\emph{14B student}} \\
    \midrule
    post-SeqKD Self (14B) & $1.559$ & -- & $0.171$ & $9.47$ & -- \\
    0.5B & $3.146$ & $+1.587$ & $0.171$ & $16.85$ & $+7.38$ \\
    1.5B & $2.119$ & $+0.561$ & $0.153$ & $16.96$ & $+7.49$ \\
    3B & $2.829$ & $+1.270$ & $0.199$ & $17.69$ & $+8.22$ \\
    vanilla 7B & $2.163$ & $+0.604$ & $0.166$ & $13.31$ & $+3.84$ \\
    vanilla Self (14B) & $2.278$ & $+0.720$ & $0.182$ & $13.47$ & $+4.00$ \\
    \midrule
    \multicolumn{6}{c}{\emph{72B student}} \\
    \midrule
    post-SeqKD Self (72B) & $1.200$ & -- & $0.200$ & $7.14$ & -- \\
    0.5B & $2.146$ & $+0.946$ & $0.171$ & $11.51$ & $+4.38$ \\
    1.5B & $1.565$ & $+0.365$ & $0.161$ & $11.93$ & $+4.79$ \\
    3B & $1.865$ & $+0.665$ & $0.191$ & $11.63$ & $+4.49$ \\
    vanilla 7B & $1.499$ & $+0.299$ & $0.168$ & $9.06$ & $+1.93$ \\
    14B & $1.599$ & $+0.399$ & $0.174$ & $9.29$ & $+2.15$ \\
    32B & $1.548$ & $+0.348$ & $0.182$ & $9.63$ & $+2.49$ \\
    vanilla Self (72B) & $1.669$ & $+0.469$ & $0.227$ & $11.81$ & $+4.67$ \\
    \bottomrule
  \end{tabular}
\end{table*}

% \begin{table*}[t]
%   \centering
%   \caption{Reject-score field at the shared 14B initialization. Norms are
%   averaged over two independent sketches. Differences are base source minus
%   Self with family-wise 95\% paired-bootstrap intervals, Bonferroni-adjusted
%   over the six pre-specified contrasts.}
%   \label{tab:reject-score-field-summary}
%   \small
%   \begin{tabular}{lrrrrr}
%     \toprule
%     Reject source & Mean-field norm & $\Delta$ field $[95\%\ \mathrm{CI}]$ &
%     Coherence & Per-token norm & $\Delta$ per-token $[95\%\ \mathrm{CI}]$ \\
%     \midrule
%     Self    & $1.63$ & -- & $0.186$ & $0.0105$ & -- \\
%     0.5B    & $3.44$ & $+1.810$ $[+1.366,+2.918]$ & $0.190$ & $0.0179$ & $+0.0073$ $[+0.0048,+0.0131]$ \\
%     1.5B    & $2.41$ & $+0.776$ $[+0.357,+1.581]$ & $0.175$ & $0.0169$ & $+0.0063$ $[+0.0039,+0.0112]$ \\
%     3B      & $2.57$ & $+0.932$ $[+0.650,+1.564]$ & $0.180$ & $0.0168$ & $+0.0063$ $[+0.0043,+0.0106]$ \\
%     7B      & $2.41$ & $+0.779$ $[+0.395,+1.449]$ & $0.192$ & $0.0152$ & $+0.0046$ $[+0.0022,+0.0082]$ \\
%     raw 14B & $2.20$ & $+0.563$ $[+0.256,+1.148]$ & $0.178$ & $0.0141$ & $+0.0036$ $[+0.0012,+0.0075]$ \\
%     72B     & $2.47$ & $+0.833$ $[+0.503,+1.201]$ & $0.225$ & $0.0215$ & $+0.0110$ $[+0.0063,+0.0180]$ \\
%     \bottomrule
%   \end{tabular}
% \end{table*}

\begin{figure*}[t]
  \centering
  \includegraphics[width=\textwidth]{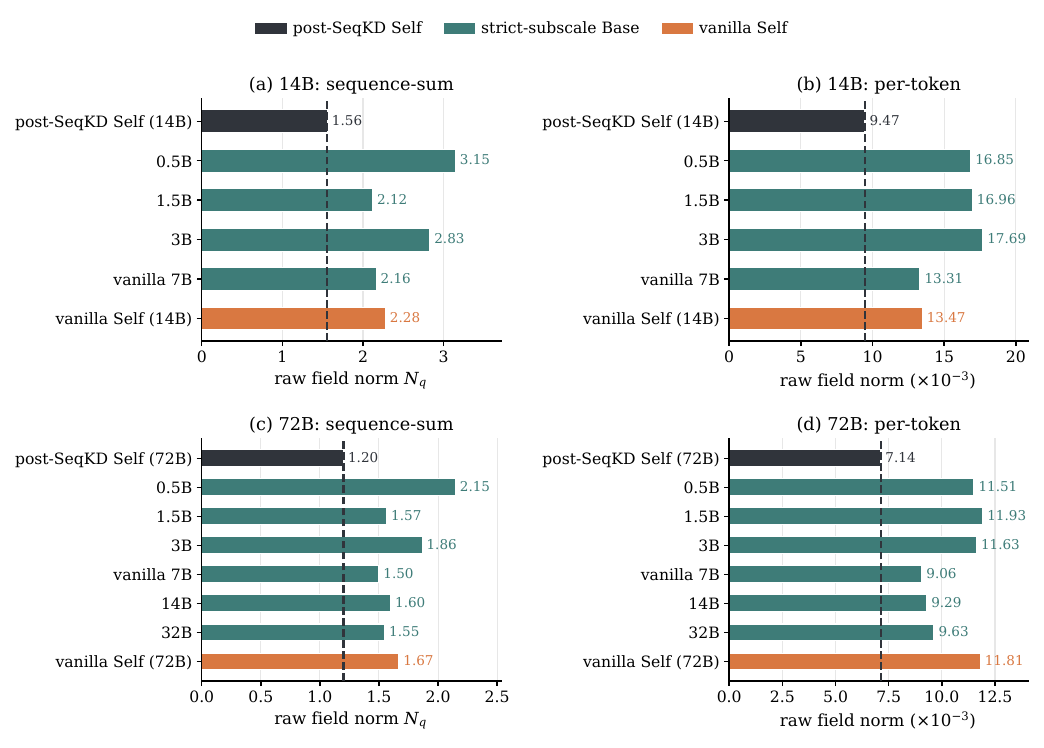}
  \caption{Empirical rejected-response fields at the shared DPO
  initialization for the 14B and 72B students. Post-SeqKD Self has an
  attenuated field relative to smaller Base sources, while vanilla Self overlaps
  the smaller Base range.}
  \label{fig:reject-state-score-field}
\end{figure*}

\begin{figure*}[t]
  \centering
  \includegraphics[width=\textwidth]{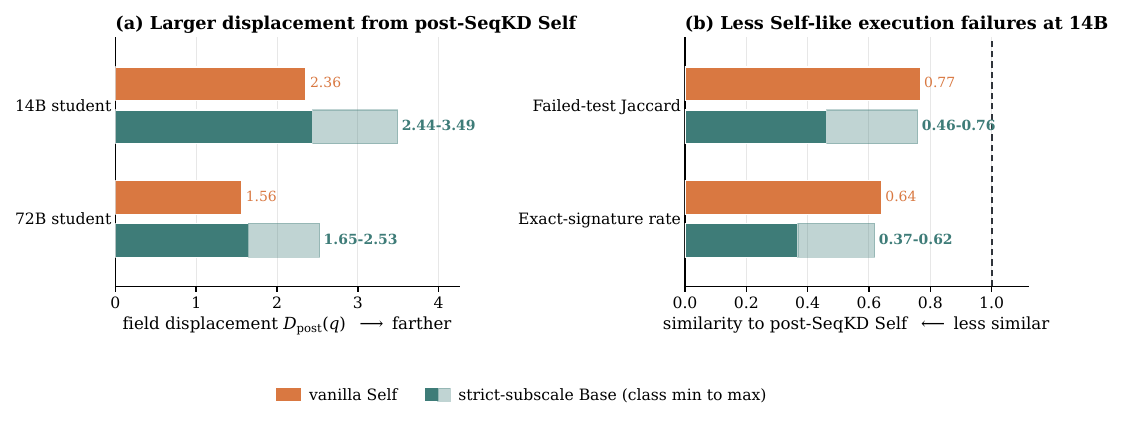}
  \caption{Differences between Self and smaller Base reject classes.
  The panels report score-field displacement from post-SeqKD Self and overlap
  in failed execution tests. These measurements characterize the source
  classes but do not rank their downstream utility.}
  \label{fig:reject-class-differences}
\end{figure*}

Post-SeqKD Self has the weakest aggregate reject field at both verified
student scales. Vanilla Self restores field strength but remains closer to
post-SeqKD Self in field space and execution failures. Smaller Base
sources combine an active field with greater displacement. Since vanilla Self
overlaps the Base field range while producing lower downstream utility, field
strength alone does not explain the source class gap.

\subsection{Fixed-Pair Dynamics and RC-DPO}

Fixed-pair displacement and Pathway III follow
\Cref{eq:fixed-pair-displacement}. Every checkpoint is
evaluated on the same bank of 512 prompts and fixed responses, so movement
across checkpoints reflects policy change rather than resampling. Endpoint
statistics and KoDCode regressions for the 14B source sweep are reported in
\Cref{tab:fixed-pair-pathway-summary}.

\begin{table*}[t]
  \centering
  \caption{Endpoint fixed-pair dynamics and KoDCode outcomes. Path III is the
  percentage of prompts that preserve the chosen response while suppressing
  the source's own reject. Regressions count problems solved by the shared
  preference-optimization initialization but not by the final model. The
  lower panel reports descriptive correlations across the six source
  endpoints.}
  \label{tab:fixed-pair-pathway-summary}
  \small
  \begin{tabular}{lrrr}
    \toprule
    Reject & Chosen $\Delta\ell$ & Path III (\%) & Regressions \\
    \midrule
    post-SeqKD Self & $-6.79$  & $5.3$  & $288$ \\
    0.5B    & $-6.67$  & $11.7$ & $262$ \\
    1.5B    & $-10.92$ & $7.8$  & $324$ \\
    3B      & $-10.19$ & $12.5$ & $247$ \\
    7B      & $-8.80$  & $12.3$ & $215$ \\
    vanilla Self & $-8.23$  & $9.6$  & $299$ \\
    % 72B     & $-12.41$ & $9.4$  & $263$ \\
    \bottomrule
  \end{tabular}

  \medskip
  \begin{tabular}{lrrr}
    \toprule
    Fixed-pair statistic
      & Gain $r/\rho$
      & Recoveries $r/\rho$
      & Regressions $r/\rho$ \\
    \midrule
    Chosen mean displacement
      & $+0.023/+0.071$ & $-0.014/+0.071$ & $-0.041/-0.107$ \\
    Chosen preservation rate
      & $+0.505/+0.643$ & $-0.106/-0.036$ & $-0.716/-0.786$ \\
    Pathway-II rate
      & $-0.511/-0.667$ & $+0.109/+0.072$ & $+0.726/+0.811$ \\
    Own-reject suppression depth
      & $+0.352/+0.107$ & $+0.292/+0.321$ & $-0.137/-0.071$ \\
    Cross-source suppression
      & $+0.360/+0.429$ & $+0.204/+0.536$ & $-0.233/-0.179$ \\
    Own-specific suppression
      & $+0.289/-0.036$ & $+0.278/+0.357$ & $-0.075/+0.071$ \\
    \bottomrule
  \end{tabular}

  \smallskip
  \begin{minipage}{0.94\textwidth}
    \footnotesize
    Here $r$ and $\rho$ denote Pearson and Spearman correlations,
    respectively. These seven-source associations describe completed runs;
    they are not prompt-level selection results.
  \end{minipage}
\end{table*}

\begin{figure*}[t]
  \centering
  \includegraphics[width=\textwidth]{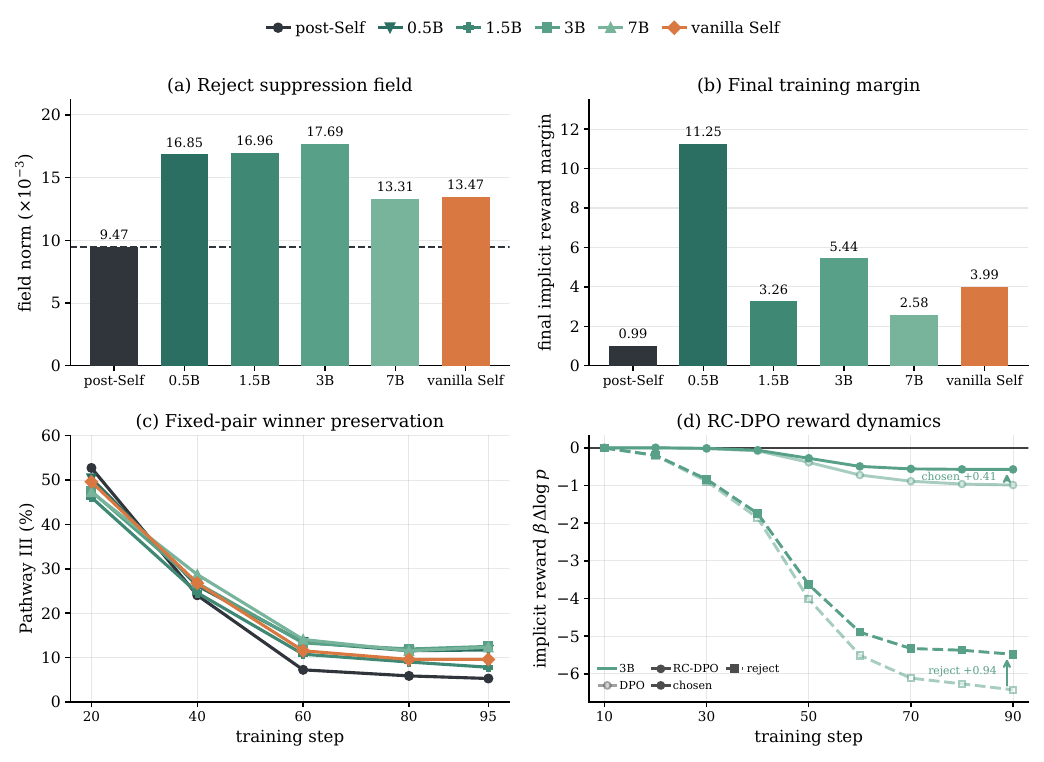}
  \caption{Training diagnostics for the 14B student. Panels report the
  initialization reject field, final implicit reward margin, winner
  preservation on fixed pairs, and the effect of RC-DPO on chosen and rejected
  rewards. None consistently separates smaller Base rejects from both
  Self controls.}
  \label{fig:kodcode-14b-dynamics-summary}
\end{figure*}

\begin{figure*}[t]
  \centering
  \includegraphics[width=\textwidth]{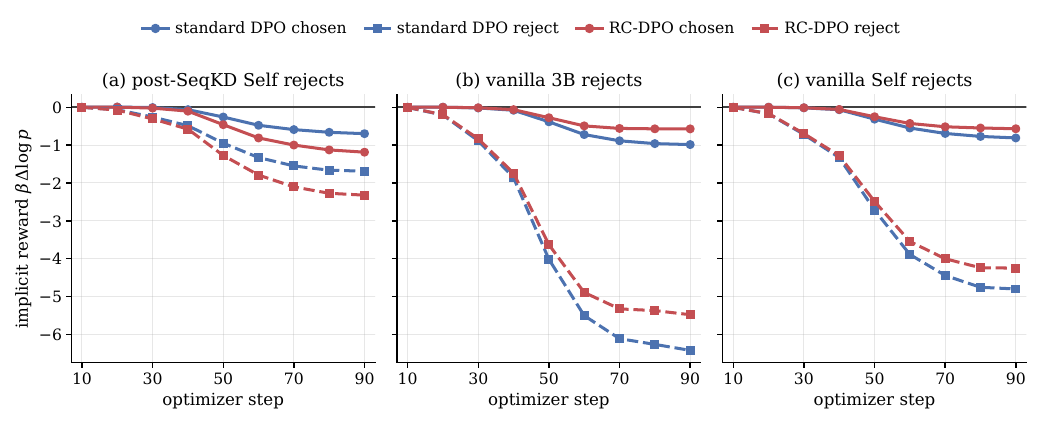}
  \caption{Reward dynamics under standard DPO and RC-DPO for the 14B student.
  Panels show post-SeqKD Self, 3B Base, and vanilla Self rejects. Solid lines
  track chosen implicit reward and dashed lines track rejected implicit reward.
  RC-DPO changes the likelihood trajectory in a source-dependent manner.}
  \label{fig:app-rcdpo-reward-dynamics}
\end{figure*}

RC-DPO directly changes the chosen and rejected gradient balance. For the 3B
Base and vanilla Self sources it increases fixed-bank Pathway III, while the
held out changes do not follow a common positive direction. The corresponding
fixed-pair effects and source by objective interactions appear in
\Cref{tab:rcdpo-source-factorial}. The implicit reward trajectories are shown
in \Cref{fig:app-rcdpo-reward-dynamics}, with absolute held-out endpoints in
\Cref{tab:rcdpo-functional-results}.

% Generated by src/analysis/plot_rcdpo_source_factorial.py.
\begin{table*}[t]
  \centering
  \caption{Reject source $\times$ objective factorial on the fixed 512-prompt bank. Likelihood displacements are sequence sums relative to the shared SeqKD reference. RC effects and interactions are percentage points.}
  \label{tab:rcdpo-source-factorial}
  \scriptsize
  \setlength{\tabcolsep}{3pt}
  \begin{tabular}{llrrrr}
    \toprule
    Reject source & Objective & Chosen $\Delta\ell$ & Own-reject $\Delta\ell$ & Path III (\%) & RC effect on Path III \\
    \midrule
    post-SeqKD Self & DPO & $-6.787$ & $-16.337$ & $5.3$ & -- \\
     & RC-DPO & $-11.650$ & $-22.545$ & $1.0$ & $-4.30$ \\
    \addlinespace
    3B Base & DPO & $-10.185$ & $-63.464$ & $12.5$ & -- \\
     & RC-DPO & $-5.890$ & $-53.974$ & $18.6$ & $+6.05$ \\
    \addlinespace
    vanilla Self & DPO & $-8.230$ & $-44.353$ & $9.6$ & -- \\
     & RC-DPO & $-5.711$ & $-39.027$ & $12.9$ & $+3.32$ \\
    \bottomrule
  \end{tabular}

  \medskip
  \begin{tabular}{lrrr}
    \toprule
    Path-III source contrast & DPO gap & RC-DPO gap & Source $\times$ RC interaction \\
    \midrule
    3B Base $-$ post-Self & $+7.23$ & $+17.58$ & $+10.35$ \\
    vanilla Self $-$ post-Self & $+4.30$ & $+11.91$ & $+7.62$ \\
    3B Base $-$ vanilla Self & $+2.93$ & $+5.66$ & $+2.73$ \\
    \bottomrule
  \end{tabular}

  \smallskip
  \begin{minipage}{0.94\textwidth}
    \footnotesize
    Every cell uses identical response IDs and token counts; no response is     truncated. All six policies suppress their own reject on 100\% of fixed     pairs. RC-DPO raises Path III for the raw sources but lowers it for Self,     so objective calibration widens rather than removes the source gap.
  \end{minipage}
\end{table*}

\begin{table*}[t]
  \centering
  \caption{Held-out performance under standard DPO and RC-DPO for the 14B student. Values are absolute percentages.}
  \label{tab:rcdpo-functional-results}
  \small
  \begin{tabular}{llrrr}
    \toprule
    Reject source & Objective & KoDCode pass@1 & BCB-C avg@4 & BCB-C pass@4 \\ 
    \midrule
    post-SeqKD Self & DPO & $63.32$ & $45.70$ & $65.35$ \\
     & RC-DPO & $62.16$ & $44.96$ & $64.65$ \\
    \addlinespace
    3B Base & DPO & $64.34$ & $51.12$ & $65.44$ \\
     & RC-DPO & $64.34$ & $50.86$ & $64.91$ \\
    \addlinespace
    vanilla Self & DPO & $61.96$ & $49.06$ & $63.95$ \\
     & RC-DPO & $62.52$ & $48.84$ & $64.12$ \\
    \bottomrule
  \end{tabular}
\end{table*}

\clearpage

\section{Reject Construction and Optimization Analyses}
\label{sec:app-theory-scope}

\subsection{Illustrative Code Example with Two Features}
\label{sec:app-theory-code-example}

Consider returning the maximum element of a nonempty integer list,
including lists whose elements are all negative.
The following constructed responses illustrate two distinct features.

\paragraph{Correct response $y_+$.}
\begin{verbatim}
def solve(xs):
    return max(xs)
\end{verbatim}

\paragraph{Incorrect response $y_e$: minimum selection.}
\begin{verbatim}
def solve(xs):
    return min(xs)
\end{verbatim}

\paragraph{Incorrect response $y_c$: an erroneous zero lower bound.}
\begin{verbatim}
def solve(xs):
    return max(0, max(xs))
\end{verbatim}

On \texttt{xs = [-3, -1]}, the outputs are $-1$, $-3$, and $0$,
respectively. Define a two-dimensional feature vector whose first
coordinate distinguishes maximum from minimum selection and whose
second coordinate marks the zero lower bound:
\begin{equation}
  \phi_+=(1,0),\qquad
  \phi_e=(-1,0),\qquad
  \phi_c=(1,1),\qquad
  \phi_0=(0,0).
  \label{eq:theory-code-features}
\end{equation}
Here $y_0$ is a task-neutral text. With
$v=(v_1,v_2)$, the change in response score is $v^\top\phi_y$.
The parameter $v_1$ favors maximum selection, while $v_2$ favors
the extra zero lower bound.

The chosen--rejected feature differences are
\begin{equation}
  \delta_e=(2,0),\qquad
  \delta_c=(0,-1),\qquad
  \delta_0=(1,0).
\end{equation}
At initialization, the corresponding single-pair DPO increments are
$\eta\beta(1,0)$, $\frac{\eta\beta}{2}(0,-1)$, and
$\frac{\eta\beta}{2}(1,0)$.
The pair involving $y_e$ reinforces maximum selection.
The pair involving $y_c$ cancels along the shared maximum-selection
coordinate and corrects the error along the second coordinate.
Thus sharing a feature and supplying a useful contrast can occur
within the same rejected response.

In this example, with $Q_0=\delta_{y_0}$ and
$\widehat u=(\widehat u_1,\widehat u_2)$, the transfer scores are
$k_e=\widehat u_1$ and
$k_c=-\widehat u_1-\widehat u_2$.
Their values depend on the utility gradient under the reference.
The two-dimensional assignment therefore separates overlap on one
feature from the response's net contribution across all features.

\subsection{Derivation of the Vector-Feature DPO Model}
\label{sec:app-vector-model}

\subsubsection{Policy Construction}
\label{sec:app-vector-policy}

Let $S(y\mid x)$ be the frozen SeqKD reference policy.
For each prompt $x$, let $\mathcal Y_x$ denote its response
support, on which $S(y\mid x)>0$.
We assign each response a fixed feature vector
$\phi(x,y)\in\mathbb R^d$ and introduce a trainable coefficient
vector $v\in\mathbb R^d$, initialized at zero.
Define the response score
\begin{equation}
  f_v(x,y)=\log S(y\mid x)+v^\top\phi(x,y).
  \label{eq:app-response-score}
\end{equation}
The score change is linear in $v$.
Normalizing the exponentiated scores gives
\begin{equation}
  \begin{aligned}
  Z_v(x)
    &=\sum_{y'\in\mathcal Y_x}
      S(y'\mid x)\exp\!\left(v^\top\phi(x,y')\right),\\
  \pi_v(y\mid x)
    &=\frac{S(y\mid x)\exp\!\left(v^\top\phi(x,y)\right)}
            {Z_v(x)}.
  \end{aligned}
  \label{eq:app-feature-policy}
\end{equation}
Thus, $\pi_v$ is a softmax over the scores $f_v$.
Equivalently,
$Z_v(x)=\mathbb E_{y'\sim S(\cdot\mid x)}
[\exp(v^\top\phi(x,y'))]$.

We assume that $Z_v(x)$ is finite in the parameter region
considered and that differentiation can be interchanged with
the sums below. Finite response support with finite features
satisfies these conditions; bounded features also ensure a
finite normalizer on countable support.

At initialization,
\begin{equation}
  Z_0(x)=\sum_y S(y\mid x)=1,
  \qquad
  \pi_0(y\mid x)=S(y\mid x).
  \label{eq:app-reference-initialization}
\end{equation}
The model therefore starts from the same reference policy as DPO.
During training, $S$ and $\phi$ remain fixed, and only $v$ changes.

\subsubsection{Connection to Local Linearization}
\label{sec:app-local-linearization}

Let $\pi_\theta$ denote the original network, with SeqKD
parameters $\theta_0$ and
$S=\pi_{\theta_0}$.
When $v$ represents displacement in the same parameter space,
the corresponding network parameters are $\theta_0+v$.
For a twice continuously differentiable log probability,
Taylor expansion around $\theta_0$ gives
\begin{equation}
  \log\pi_{\theta_0+v}(y\mid x)
  =
  \log S(y\mid x)
  +v^\top g(x,y)
  +O(\|v\|^2),
  \label{eq:app-network-taylor}
\end{equation}
where
\begin{equation}
  g(x,y)
  =
  \left.
  \nabla_\theta\log\pi_\theta(y\mid x)
  \right|_{\theta=\theta_0}.
  \label{eq:app-initial-score-gradient}
\end{equation}
The expansion is pointwise in $(x,y)$; a uniform remainder
requires corresponding uniform derivative bounds.

Choosing $\phi=g$ uses the network's initial score gradients
as fixed response features. Exponentiating the linear part
of \Cref{eq:app-network-taylor} gives the unnormalized weights
$S(y\mid x)\exp(v^\top g(x,y))$.
Normalizing these weights yields
\Cref{eq:app-feature-policy}.
This construction uses initial gradient features throughout
training, following the local linearization perspective
on neural training~\citep{linearizedNetworks}.

To verify its first-order correspondence with the network,
first differentiate the log normalizer:
\begin{equation}
  \begin{aligned}
  \nabla_v\log Z_v(x)
  &=
  \frac{
    \sum_{y'}S(y'\mid x)
    \exp(v^\top\phi(x,y'))\phi(x,y')
  }{Z_v(x)}\\
  &=
  \mathbb E_{y'\sim\pi_v(\cdot\mid x)}
  [\phi(x,y')].
  \end{aligned}
  \label{eq:app-normalizer-gradient}
\end{equation}
Consequently,
\begin{equation}
  \nabla_v\log\pi_v(y\mid x)
  =
  \phi(x,y)
  -
  \mathbb E_{y'\sim\pi_v(\cdot\mid x)}
  [\phi(x,y')].
  \label{eq:app-policy-score-gradient}
\end{equation}

For the choice $\phi=g$, the reference-weighted feature
mean is zero:
\begin{equation}
  \begin{aligned}
  \mathbb E_{y\sim S(\cdot\mid x)}[g(x,y)]
  &=
  \sum_y \pi_{\theta_0}(y\mid x)
  \left.
  \nabla_\theta\log\pi_\theta(y\mid x)
  \right|_{\theta=\theta_0}\\
  &=
  \sum_y
  \left.
  \nabla_\theta\pi_\theta(y\mid x)
  \right|_{\theta=\theta_0}\\
  &=
  \left.
  \nabla_\theta\sum_y\pi_\theta(y\mid x)
  \right|_{\theta=\theta_0}
  =0.
  \end{aligned}
  \label{eq:app-zero-mean-score}
\end{equation}
Here the response support is fixed locally in $\theta$,
and the interchange of differentiation and summation
is assumed valid.
Combining \Cref{eq:app-policy-score-gradient} with
\Cref{eq:app-zero-mean-score} gives
\begin{equation}
  \left.
  \nabla_v\log\pi_v(y\mid x)
  \right|_{v=0}
  =
  g(x,y)
  =
  \left.
  \nabla_\theta\log\pi_\theta(y\mid x)
  \right|_{\theta=\theta_0}.
  \label{eq:app-first-order-matching}
\end{equation}
Thus the constructed policy matches both the network's
response probabilities and its log-probability gradients
at initialization.

The modeling assumption is that these response features
remain fixed as $v$ is trained. The subsequent derivations
are exact for this fixed-feature family. Its approximation
to the original network beyond initialization depends on
the higher-order terms in
\Cref{eq:app-network-taylor}.

\subsubsection{Shared Features and Response Contrasts}
\label{sec:app-vector-contrasts}

For a chosen response $y^+$ and a reject $y^-$ to the same
prompt, define
\begin{equation}
  \delta(x,y^+,y^-)
  =
  \phi(x,y^+)-\phi(x,y^-).
  \label{eq:app-feature-contrast}
\end{equation}
Their log-probability ratio satisfies
\begin{equation}
  \log\frac{\pi_v(y^+\mid x)}{\pi_v(y^-\mid x)}
  =
  \log\frac{S(y^+\mid x)}{S(y^-\mid x)}
  +v^\top\delta.
  \label{eq:app-pair-log-ratio}
\end{equation}
If the two responses have identical feature values in
coordinate $j$, then $\delta_j=0$, and changing $v_j$
does not change their probability ratio.
If their feature values differ, changing $v_j$ changes
the ratio according to $\delta_j$.
A pair can therefore share features in some coordinates
and contrast in others.

For example, the illustrative assignment
$\phi(x,y^+)=(1,1)$ and $\phi(x,y^-)=(1,-1)$
gives $\delta=(0,2)$.
The first coordinate changes both response scores equally,
whereas the second changes their relative score.
Such assignments illustrate the model's geometry.
When $\phi$ consists of network score gradients, its
coordinates instead represent sensitivities to network
parameters. \Cref{sec:app-theory-code-example} gives a
concrete code example with two features.

\subsubsection{DPO Margin and Gradient}
\label{sec:app-vector-dpo}

For a preference triple $(x,y^+,y^-)$, define the
reference-relative DPO margin
\begin{equation}
  m_v
  =
  \beta\left[
    \log\frac{\pi_v(y^+\mid x)}{S(y^+\mid x)}
    -
    \log\frac{\pi_v(y^-\mid x)}{S(y^-\mid x)}
  \right],
  \qquad \beta>0.
  \label{eq:app-dpo-margin-definition}
\end{equation}
By \Cref{eq:app-feature-policy},
\begin{equation}
  \log\frac{\pi_v(y\mid x)}{S(y\mid x)}
  =
  v^\top\phi(x,y)-\log Z_v(x).
  \label{eq:app-reference-log-ratio}
\end{equation}
The chosen and rejected responses share the same prompt,
so their normalizers cancel:
\begin{equation}
  \begin{aligned}
  m_v
  &=
  \beta\left[
    v^\top\phi(x,y^+)-\log Z_v(x)
    -v^\top\phi(x,y^-)+\log Z_v(x)
  \right]\\
  &=\beta v^\top\delta.
  \end{aligned}
  \label{eq:app-linear-dpo-margin}
\end{equation}
A positive margin means that the chosen-to-rejected
probability ratio has increased relative to the reference.

The pairwise DPO loss is
\begin{equation}
  \ell(v;x,y^+,y^-)
  =
  -\log\sigma(m_v),
  \qquad
  \sigma(z)=\frac{1}{1+e^{-z}}.
  \label{eq:app-pairwise-dpo-loss}
\end{equation}
Using
\begin{equation}
  \frac{d}{dm}\big[-\log\sigma(m)\big]
  =-\sigma(-m),
  \qquad
  \nabla_v m_v=\beta\delta,
  \label{eq:app-dpo-chain-rule}
\end{equation}
the pairwise gradient is
\begin{equation}
  \nabla_v\ell(v;x,y^+,y^-)
  =
  -\beta\sigma(-\beta v^\top\delta)\delta.
  \label{eq:app-pairwise-dpo-gradient}
\end{equation}
Hence a gradient-descent step on this pair changes $v$ by
\begin{equation}
  -\eta\nabla_v\ell
  =
  \eta\beta\sigma(-\beta v^\top\delta)\delta.
  \label{eq:app-pairwise-dpo-update}
\end{equation}
The update follows the feature difference $\delta$.
Its scalar weight decreases continuously as the margin
increases, since
\begin{equation}
  \frac{d}{dm}\sigma(-m)
  =
  -\sigma(-m)\sigma(m)<0.
  \label{eq:app-margin-weight}
\end{equation}
At initialization, every pair has zero margin and
weight $\sigma(0)=1/2$.

\subsubsection{The Reject-Dependent Update Field}
\label{sec:app-vector-field}

Using the preference-prompt law $p$, the verified chosen-response
law $T$, and the reject law $Q$ induced by the construction, define
the preference law
\begin{equation}
  \mathcal D_Q(x,y^+,y^-)
  =
  p(x)T(y^+\mid x)Q(y^-\mid x),
  \label{eq:app-preference-law}
\end{equation}
and write $\mathbb E_Q$ for expectation under this law.
The population DPO objective is
\begin{equation}
  \mathcal L_Q(v)
  =
  \mathbb E_Q[\ell(v;x,y^+,y^-)],
  \label{eq:app-population-dpo-loss}
\end{equation}
where the response feature contrast $\delta$ depends on
the sampled triple as in \Cref{eq:app-feature-contrast}.
Assuming differentiation and expectation can be interchanged,
\Cref{eq:app-pairwise-dpo-gradient} gives
\begin{equation}
  \nabla_v\mathcal L_Q(v)
  =
  -\beta\mathbb E_Q
  \left[\sigma(-\beta v^\top\delta)\delta\right].
  \label{eq:app-population-dpo-gradient}
\end{equation}
Define the negative-gradient field
\begin{equation}
  F_Q(v)
  =
  -\nabla_v\mathcal L_Q(v)
  =
  \beta\mathbb E_Q
  \left[\sigma(-\beta v^\top\delta)\delta\right].
  \label{eq:app-update-field}
\end{equation}
Full-batch gradient descent therefore follows
\begin{equation}
  v_{t+1}^Q
  =
  v_t^Q+\eta_tF_Q(v_t^Q),
  \qquad
  v_0^Q=0,
  \label{eq:app-full-batch-dynamics}
\end{equation}
with positive step sizes $\eta_t$.
For a fixed finite preference dataset, replacing
$\mathbb E_Q$ by the empirical average yields the
corresponding empirical objective and exact full-batch update.

At initialization,
\begin{equation}
  F_Q(0)
  =
  \frac{\beta}{2}\mathbb E_Q[\delta].
  \label{eq:app-initial-update-field}
\end{equation}
For a comparison reject distribution $Q_0$, define
\begin{equation}
  \bar\phi_Q(x)
  =
  \mathbb E_{y^-\sim Q(\cdot\mid x)}[\phi(x,y^-)],
  \qquad
  \bar\phi_0(x)
  =
  \mathbb E_{y^-\sim A(\cdot\mid x)}[\phi(x,y^-)].
\end{equation}
Because the prompt and chosen distributions are shared,
their contributions cancel in the initial field difference:
\begin{equation}
  F_Q(0)-F_{Q_0}(0)
  =
  \frac{\beta}{2}
  \mathbb E_{x\sim p}
  \left[\bar\phi_0(x)-\bar\phi_Q(x)\right].
  \label{eq:app-initial-source-difference}
\end{equation}
This identity connects reject construction to the initial
update used in the transfer analysis.

More generally, with $p$, $T$, and $\phi$ fixed,
reject construction affects $F_Q(v)$ through the induced
distribution of $\delta$.
At initialization, only its mean enters the update.
As training proceeds, the margin-dependent weights make
the update depend on its broader distribution.

\subsection{Vector DPO Dynamics and the Finite-Horizon Bound}
\label{sec:app-theory-proofs}

\paragraph{Policy family and feature gradients.}
Use the policy and update field derived in
\Cref{sec:app-vector-model}, with the stated regularity conditions.
Write $g_v(x,y)=\nabla_v\log\pi_v(y\mid x)$, whose centered-feature
expression is given in \Cref{eq:app-policy-score-gradient}.
For the bounds below, assume finite response sets and bounded features,
or response spaces on which the corresponding uniform bounds exist.

\paragraph{Bounds on the complete vector field.}
The field in \Cref{eq:theory-feature-field} satisfies
\begin{align}
  \|F_Q(v)\|&\le M_Q:=\beta\mathbb E_Q\|\delta\|,\\
  \nabla F_Q(v)&=
  -\beta^2\mathbb E_Q\left[
  \sigma(-\beta v^\top\delta)\sigma(\beta v^\top\delta)
  \delta\delta^\top\right].
\end{align}
In particular, a global Lipschitz constant is
\begin{equation}
  L_Q=\frac{\beta^2}{4}
      \left\|\mathbb E_Q[\delta\delta^\top]\right\|_{\mathrm{op}}.
  \label{eq:theory-feature-second-moment}
\end{equation}
This matrix retains interactions among all feature coordinates.
For $\|\delta\|\le D$, common bounds over all compared sources are
$M=\beta D$ and $L=\beta^2D^2/4$.
For a fixed source family, the maxima of $M_Q$ and $L_Q$ give
tighter common constants.

For the utility $U$ defined in \Cref{sec:theory-region},
\begin{equation}
  u=\mathbb E_{x\sim p_{\mathrm{eval}}}
       \operatorname{Cov}_{y\sim S(\cdot\mid x)}
       \bigl(r(x,y),\phi(x,y)\bigr).
  \label{eq:theory-utility-gradient}
\end{equation}
If $\|\phi\|\le B$, then $\|g_v\|\le2B$ and
\begin{equation}
  \nabla^2 U(v)=
  \mathbb E_{x\sim p_{\mathrm{eval}},y\sim\pi_v}
  \left[(r(x,y)-U_x(v))g_v(x,y)g_v(x,y)^\top\right],
\end{equation}
where $U_x(v)=\mathbb E_{\pi_v(\cdot\mid x)}r(x,y)$.
Hence $L_U=4B^2$ is a valid global bound for rewards in $[0,1]$.
A smaller local curvature bound can be used on the trajectory ball.

\paragraph{Formal conditions.}
We prove \Cref{prop:theory-gain} in the following form: the compared
constructions $Q$ and $Q_0$ share the constants $M$, $L$, and $L_U$ above,
$u=\nabla U(0)\ne0$, and \Cref{eq:app-full-batch-dynamics} runs for $H$
steps with step sizes $\eta_t>0$ and total step size
$\tau_H=\sum_{t=0}^{H-1}\eta_t$. The conclusion is
\Cref{eq:theory-gain-bound} with $E_H$ given explicitly in
\Cref{eq:app-error-constant}.

\paragraph{Step 1: extract initial transfer.}
Taking the inner product of \Cref{eq:app-initial-source-difference}
with $u$ and using the transfer score $k$ of
\Cref{eq:theory-transfer-score} gives the exact identity
\begin{equation}
  \left\langle u,F_Q(0)-F_{Q_0}(0)\right\rangle
  =\frac{\beta}{2}\|u\|\kappa_Q.
  \label{eq:theory-field-comparison}
\end{equation}
Splitting $k$ into positive and negative parts, with expectations over
$x\sim p$ and $y\sim Q$,
\begin{equation}
  \kappa_Q=\mathbb E[k]_+-\mathbb E[-k]_+,\qquad
  a_Q=\mathbb E[k]_++\mathbb E[-k]_+,\qquad
  a_Qc_Q=\mathbb E[-k]_+,
\end{equation}
so $\kappa_Q=a_Q-2a_Qc_Q=a_Q(1-2c_Q)$.

\paragraph{Step 2: control vector trajectory drift.}
Let $\tau_t=\sum_{j<t}\eta_j$.
Since $\|F_Q\|\le M$, all iterates obey $\|v_t^Q\|\le M\tau_t$.
Writing $R_Q=v_H^Q-\tau_HF_Q(0)$, Lipschitz continuity gives
\begin{align}
  \|R_Q\|
  &\le\sum_{t=0}^{H-1}\eta_t
       \|F_Q(v_t^Q)-F_Q(0)\|\\
  &\le LM\sum_{t=0}^{H-1}\eta_t \tau_t
   =\frac{LM}{2}\left(\tau_H^2-\sum_{t=0}^{H-1}\eta_t^2\right).
\end{align}
The same bound holds for $Q_0$.
It controls the full vector remainder, including directions
orthogonal to the initial utility gradient.

\paragraph{Step 3: compare endpoint utilities.}
Taylor's inequality on the trajectory ball gives
\begin{equation}
  |U(v)-U(0)-u^\top v|\le\frac{L_U}{2}\|v\|^2.
\end{equation}
Applying the lower inequality to $v_H^Q$ and the upper inequality to
$v_H^{Q_0}$ yields
\begin{align}
  G_H(Q)
  &\ge u^\top(v_H^Q-v_H^{Q_0})
       -\frac{L_U}{2}
        \left(\|v_H^Q\|^2+\|v_H^{Q_0}\|^2\right)\\
  &\ge \tau_Hu^\top(F_Q(0)-F_{Q_0}(0))
       -\|u\|(\|R_Q\|+\|R_{Q_0}\|)-L_UM^2\tau_H^2.
\end{align}
Substituting the bounds on $\|R_Q\|$ and $\|R_{Q_0}\|$ and collecting
the correction terms proves \Cref{prop:theory-gain} with
\begin{equation}
  E_H=\|u\|LM\left(\tau_H^2-\sum_{t=0}^{H-1}\eta_t^2\right)
      +L_UM^2\tau_H^2.
  \label{eq:app-error-constant}
\end{equation}
The correction is a worst case over the compared constructions and does
not vanish as $Q\to Q_0$.
If a source has $\kappa_Q>0$ and the schedule is scaled by
$\lambda>0$, then $K_H=\lambda K_*$ and $E_H=\lambda^2 E_*$.
For sufficiently small $\lambda$, the bound is positive.
Any target $\epsilon$ up to that positive bound makes
\Cref{eq:theory-favorable-region} nonempty.The coordinates $a_Q$ and $c_Q$ are evaluated at the DPO initialization, and
the bound is informative only while $\tau_H$ is small enough that $E_H$ does
not dominate. The leading transfer term is ordered by $\kappa_Q$, and the bound guarantees a positive utility gain whenever $K_H\kappa_Q>E_H$. The intervention identities in \S F.4 motivate P1--P3 through their effects on this transfer term.

\subsection{Intervention Identities and Implementation}
\label{sec:app-theory-selection}

The interventions instantiate the comparison distribution $Q_0$ of
\Cref{sec:theory-region} as follows: randomized mixtures compare against
the replaced Self source, the structural controls compare against
length-matched gibberish, and reference-likelihood reselection compares
against native selection ($\gamma=0$).

\paragraph{Source composition.}
Since $k$ is fixed by the common reference, utility, and comparison
distribution, randomized source assignment gives
\begin{equation}
  \kappa_{\mathcal M_{\mathbf w}}=\sum_s w_s\kappa_{Q_s},
  \qquad
  a_{\mathcal M_{\mathbf w}}=\sum_s w_sa_{Q_s},
  \qquad
  a_{\mathcal M_{\mathbf w}}c_{\mathcal M_{\mathbf w}}
      =\sum_s w_sa_{Q_s}c_{Q_s}.
\end{equation}
The adverse share mixes with transfer-strength weights.
Uniform $M,L,L_U$ across the source family make the gain lower bound
affine in source weights. The actual endpoint includes the
source-dependent trajectory and curvature terms bounded in the proof.

\paragraph{Prompt reassignment.}
For an additive decomposition
$\phi(x,y)=\phi_{\mathrm{dom}}(y)+\phi_{\mathrm{pair}}(x,y)$,
independent reassignment preserves
$\mathbb E_Q\phi_{\mathrm{dom}}(y)$ and its contribution to
$\kappa_Q$. The prompt-dependent component can change. 
An order-invariant lexical component of $\phi_{\mathrm{dom}}$
is also preserved by lexical permutation of its token inventory.
Whether this component supplies favorable transfer is the
construction hypothesis examined by the structural controls.

\paragraph{Reference likelihood selection.}
The standardized candidate scores are
\begin{equation}
  s_j=\frac{1}{|y_j|}\log S(y_j\mid x),\qquad
  \tilde s_j=\frac{s_j-\overline s_{C_x}}{\operatorname{sd}_{C_x}(s)},
  \label{eq:theory-reference-score}
\end{equation}
and reselection samples $q_\gamma$ as in
\Cref{eq:theory-reference-tilt}.
Let $k_j=k(x,y_j)$ and hold $Q_0$, $u$, and each candidate bank fixed.
Differentiating \Cref{eq:theory-reference-tilt} gives
\begin{equation}
  \frac{d\kappa_{Q_\gamma}}{d\gamma}
  =-\mathbb E_{x\sim p}
       \operatorname{Cov}_{j\sim q_\gamma(\cdot\mid C_x)}(\tilde s_j,k_j).
  \label{eq:theory-selection-transfer}
\end{equation}
A negative averaged covariance increases the leading transfer term
as repulsion strengthens. This is the explicit connection condition
between the observed reference score and the vector model's
utility-aligned geometry.

Positive $\gamma$ favors lower-reference-likelihood candidates,
negative $\gamma$ favors higher-likelihood candidates, and
$\gamma=0$ gives native selection.
Shared Gumbel variables couple sampling randomness across arms.
The binary-wrong sweep retains failing candidates with their observed
partial test-pass fractions. 

\subsection{Connection to Gradient Interference}

At initialization, write $g^+=g_0(x,y^+)$ and $g^-=g_0(x,y^-)$.
For one pair,
\begin{equation}
  \Delta v=\frac{\eta\beta}{2}(g^+-g^-)
          =\frac{\eta\beta}{2}\delta.
  \label{eq:theory-initial-update}
\end{equation}
The first-order change in the chosen log probability is
$\frac{\eta\beta}{2}
(\|g^+\|^2-\langle g^+,g^-\rangle)$, as in gradient-interference
analyses~\citep{gradientEntanglement,likelihoodDisplacement}.
The utility gradient in \Cref{eq:theory-utility-gradient} instead
aggregates reward-relevant score directions over evaluation prompts.
The transfer score projects the change in negative supervision
onto this aggregate direction, connecting training constructions
to evaluation utility.

\end{document}